\pdfoutput=1 

\documentclass[twoside,11pt]{article}

\usepackage{jmlr2e}
\usepackage{amsfonts}
\usepackage{amsmath}
\usepackage{mathtools}
\usepackage{xcolor}
\usepackage{float}
\usepackage{booktabs}

\newcommand{\x}{\textbf{x}}

\newcommand{\f}{\textbf{f}}

\newcommand{\linop}[1]{\mathcal{L}_{#1}}
\newcommand{\sublinop}[1]{\mathcal{F}_{#1}}
\newcommand{\sgn}[1]{\text{sgn} (#1)}
\newcommand{\atan}[1]{\text{tan}^{-1} (#1)}
\newcommand{\NVI}{S} 
\newcommand*{\QEDS}{\hfill\BlackBox}

\newtheorem{alg}[theorem]{Algorithm}

\DeclareMathOperator*{\argmin}{arg\,min}

\firstpageno{1}
\usepackage{lastpage}
\jmlrheading{20}{2019}{1-\pageref{LastPage}}{1/19; Revised
7/19}{8/19}{19-065}{Christian Agrell}
\ShortHeadings{GPs with Linear Inequality Constraints}{Agrell}

\begin{document}

\setlength{\parskip}{0pt} 

\title{Gaussian Processes with Linear Operator Inequality Constraints}

\author{\name Christian Agrell \email chrisagr@math.uio.no \\
       \addr Department of Mathematics\\
       University of Oslo\\
       P.O. Box 1053 Blindern, Oslo N-0316, Norway\\
       \rule{2cm}{0.4pt} \\
       \addr Group Technology and Research \\
       DNV GL \\
       P.O. Box 300, 1322 H{{\o}}vik, Norway
       }

\editor{Andreas Krause}
\maketitle
\begin{abstract}
    This paper presents an approach for constrained Gaussian Process (GP) regression where we assume 
    that a set of linear transformations of the process are bounded.     
    It is motivated by machine learning applications for high-consequence engineering systems, 
    where this kind of information is often made available from phenomenological knowledge.
    We consider a GP $f$ over functions on $\mathcal{X} \subset \mathbb{R}^{n}$ taking values in $\mathbb{R}$, 
    where the process $\linop{}f$ is still Gaussian when $\linop{}$ is a linear operator. 
    Our goal is to model $f$ under the constraint that realizations of $\linop{}f$ are confined to a 
    convex set of functions. In particular, we require that $a \leq \linop{}f \leq b$, given two functions 
    $a$ and $b$ where $a < b$ pointwise. This formulation provides a consistent way of encoding multiple 
    linear constraints, such as shape-constraints based on e.g. boundedness, monotonicity or convexity. 
    We adopt the approach of using a sufficiently dense set of virtual observation locations 
    where the constraint is required to hold, and derive the exact posterior for a conjugate likelihood. 
    The results needed for stable numerical implementation are derived, together with an efficient sampling scheme for estimating the posterior 
    process.
\end{abstract}


\section{Introduction}
Gaussian Processes (GPs) are a flexible tool for Bayesian nonparametric function estimation, and 
widely used for applications that require inference on functions such as regression and classification. 
A useful property of GPs is that they automatically produce estimates on prediction uncertainty,
and it is often possible to encode prior knowledge in a principled manner in the modelling of prior covariance.
Some early well-known applications of GPs are within spatial statistics, e.g. meteorology \citep{Thompson:1956:Opt_smooth}, and in geostatistics \citep{Matheron:1973:IRF} 
where it is known as \emph{kriging}. 
More recently, GPs have become a popular choice within probabilistic machine learning \citep{Rasmussen:2005:GPML, Ghahramani:2015:probML}. 
Since the GPs can act as interpolators when observations are noiseless, GPs have also become the main approach
for uncertainty quantification and analysis involving computer experiments \citep{Sacks:1989:DOE, Kennedy:2001:Cal}.

Often, the modeler performing function estimation has prior knowledge, 
or at least hypotheses, on some properties of the function to be estimated. 
This is typically related to the function shape with respect to some of the input parameters, 
such as boundedness, monotonicity or convexity. 
Various methods have been proposed for imposing these types of constraints on GPs 
(see Section \ref{sec:related_work} for a short review).
For engineering and physics based applications, constraints based on 
integral operators and partial differential equations are also relevant
\citep{Jidling:20017:CGP_NIPS, Sarkka:2011:SPDE}.
What the above constraints have in common is that they are 
linear operators, and so any combination of such constraints can be written as a single linear operator. 
For instance, the constraints $a_{1}(\x) \leq f(\x) \leq b_{1}(\x) $, $ \partial f / \partial x_{i} \leq 0 $ and $\partial^{2}f / \partial x^{2}_{j} \geq 0 $
for some function (or distribution over functions) $f : X \rightarrow Y$, can be written as $a(\x) \leq \linop{} f(\x) \leq b(\x) $
for $a(\x) = [a_{1}(\x), - \infty, 0]$, $b(\x) = [b_{1}(\x), 0, \infty]$
and $\linop{} : Y^{X} \rightarrow (Y^{X})^{3}$ being the linear operator $\linop{}f = [f, \partial f / \partial x_{i}, \partial^{2}f / \partial x^{2}_{j}]$.

The motivation for including constraints is usually to improve predictions
and to obtain a reduced and more realistic estimate on the uncertainty, the latter having significant impact 
for risk-based applications.
For many real-world systems, information related to constraints in this form is often available from 
phenomenological knowledge. For engineering systems, this is typically knowledge related to some underlying 
physical phenomenon. Being able to make use of these constraint in probabilistic modelling is particularly 
relevant for high-consequence applications, where obtaining realistic uncertainty estimates in subsets of the 
domain where data is scarce is a challenge. 
Furthermore, information on whether these types of constraints are likely to hold given a set of observations 
is also useful for explainability and model falsification. 
For a broader discussion see \citep{Agrell:2018:pitfallml, Eldevik:2018:AI_Safety}. 

In this paper, we present a model for estimating a function $\textnormal{f} : \mathbb{R}^{n_{x}} \rightarrow \mathbb{R}$ by a
constrained GP (CGP) $f | D, a(\x) \leq \linop{}f(\x) \leq b(\x)$. Here 
$D$ is a set of observations of $(\x_{j}, y_{j})$, possibly including additive white noise, and
$f \sim \mathcal{GP}(\mu(\x), K(\x, \x'))$ is a GP with 
mean $\mu(\x)$ and covariance function $K(\x, \x')$ that are chosen such that existence of $\linop{}f$ is ensured.
Due to the linearity of $\linop{}$, both $\linop{}f | D$ and  $f | D, \linop{}f$ remain Gaussian, and our approach is 
based on modelling $f | D, \linop{}f$ under the constraint $a(\x) \leq \linop{}f(\x) \leq b(\x)$.
To model the constraint that $a(\x) \leq \linop{}f(\x) \leq b(\x)$ for all inputs $\x$, 
we take the approach of using a finite set of input locations where the constraint is required to hold. 
That is, we require that $a(\x_{v}) \leq \linop{}f(\x_{v}) \leq b(\x_{v})$ 
for a finite set of inputs $\{ \x_{v} \}$ called the set of \emph{virtual observation locations}. 
With this approach the CGP is not guaranteed to satisfy the constraint on
the entire domain, but a finite set of points $\{ \x_{v} \}$ can be found so that the constraint holds globally with sufficiently high probability.

The model presented in this paper is inspired by the research on shape-constrained GPs, in particular 
\citep{WangBerger:2016:CGP, DaVeiga:2012:CGP, DaVeiga:2017:CGP, Riihimaki:2010:CGP, Golchi:2015:CGP, Maatouk:2017:CGP, LpezLopera:2017:CGP}.
We refer to Section \ref{sec:monoconstr} for further discussion on these alternatives. 
In the case where $\linop{} = \partial / \partial x_{i}$, our approach is most similar to 
that of \citet{WangBerger:2016:CGP}, where the authors make use of a similar sampling scheme for noiseless GP regression applied to computer code emulation. 
Many of the approaches to constrained GPs, including ours, rely on the constraint to be satisfied at a specified set of 
virtual locations.
The use of virtual constraint observations may seem \emph{ad hoc} at first, as the set of virtual observation locations
has to be dense enough to ensure that the constraint holds globally with sufficiently high probability. 
Inversion of the covariance matrix of the joint GP may therefore be of concern, both because this scales with the number of observations cubed
and because there is typically high serial correlation if there are many virtual observations close together. 
The general solution is then to restrict the virtual observation set to regions where the probability of occurrence of
the constraint is low \citep{Riihimaki:2010:CGP, WangBerger:2016:CGP}. 
According to \citet{WangBerger:2016:CGP}, when they followed this approach in their experiments, they found 
that only a modest number of virtual observations were typically needed, that these points were usually
rather disperse, and the resulting serial correlation was not severe. We draw the same conclusion in our experiments. 
There is also one benefit with the virtual observation approach, which is that implementation of constraints that only hold on 
subsets of the domain is straightforward. 

For practical use of the model presented in this paper, we also pay special attention to numerical implementation. 
The computations involving only real observations or only virtual observations are separated, which is convenient 
when only changes to the constraints are made such as in algorithms for finding a sparse set of virtual observation locations 
or for testing/validation of constraints. We also provide the algorithms based on Cholesky factorization for stable numerical implementation, 
and an efficient sampling scheme for estimating the posterior process.
These algorithms are based on derivation of the exact posterior of the constrained Gaussian process 
using a general linear operator, and constitutes the main contribution of this paper.

The paper is structured as follows: 
In Section \ref{sec:GP_intro} we state the results needed on GP regression and GPs under linear transformations. 
Our main results are given in Section \ref{sec:CGP}, where we introduce the constrained GP (CGP) and present the model for GP regression under linear 
inequality constraints. In particular, given some training data, we derive the posterior predictive distribution of the CGP evaluated at a 
finite set of inputs, which is a compound Gaussian with a truncated Gaussian mean (Section \ref{sec:CGP_posterior}). 
Section \ref{sec:CGP_sampling} presents an algorithm for sampling from the posterior, and parameter estimation is addressed in Section \ref{sec:CGP_MLE}.
Section \ref{sec:find_xv} and Section \ref{sec:separate_obs} are dedicated to optimization of the set of virtual observation locations needed 
to ensure that the constraint holds with sufficiently high probability. 
Some relevant alternative approaches from the literature on GP's under linear ¨
constraints are discussed in Section \ref{sec:monoconstr}, 
followed up by 
numerical examples considering monotonicity and boundedness constraints.
A Python implementation is available at \url{https://github.com/cagrell/gp_constr},
together with the code used for the examples.
We end with some concluding remarks in Section \ref{sec:discussion}. 

\newpage

\section{Gaussian Processes and Linear Operators}
\label{sec:GP_intro}
We are interested in GP regression on functions $\textnormal{f} : \mathbb{R}^{n_{x}} \rightarrow \mathbb{R}$
under the additional inequality constraint $a(\x) \leq \linop{}\textnormal{f}(\x) \leq b(\x)$ for some
specified functions $a(\x)$ and $b(\x)$, and the class of linear operators $\{ \linop{} | \linop{}\textnormal{f}: \mathbb{R}^{n_{x}} \rightarrow \mathbb{R}^{n_{c}} \}$.
Here $n_{x}$ and $n_{c}$ are positive integers, and the subscripts are just used to indicate the relevant underlying space over $\mathbb{R}$.
We will make use of the properties of GPs under linear transformations given below.  

\subsection{Gaussian Process Regression}
\label{sec:GPR}
We consider a Gaussian process $f \sim \mathcal{GP}(\mu(\x), K(\x, \x'))$ given as a prior over functions $\textnormal{f} : \mathbb{R}^{n_{x}} \rightarrow \mathbb{R}$,
which is specified by its mean and covariance function 
\begin{equation}
    \label{eq:gp_prior}
    \begin{split}
        \mu(\x) &= \mathbb{E}[f(\x)] : \mathbb{R}^{n_{x}} \rightarrow \mathbb{R}, \\
        K(\x, \x') &= \mathbb{E}[(f(\x) - \mu(\x))(f(\x') - \mu(\x'))] : \mathbb{R}^{n_{x} \times n_{x}} \rightarrow \mathbb{R}.
    \end{split}
\end{equation}

Let $\x$ denote a vector in $\mathbb{R}^{n_{x}}$ and $X$ the $N \times n_{x}$ matrix of $N$ such input vectors. 
The distribution over the vector $\f$ of $N$ latent values corresponding to $X$ is then multivariate Gaussian with
\begin{equation*}
    \f | X \sim \mathcal{N}(\mu(X), K(X, X)),
\end{equation*}

where $K(X, X')$ denotes the Gram matrix $K(X, X')_{i, j} = K( \x_{i}, \x_{j}')$
for two matrices of input vectors $X$ and $X'$. 
Given a set of observations $Y = [y_{1}, \dots, y_{N}]^{T}$, and under the assumption 
that the relationship between the latent function values and observed output is Gaussian, $Y | \f \sim \mathcal{N}(\f, \sigma^{2} I_{N})$, 
the predictive distribution for new observations $X^{*}$ is still Gaussian with mean and covariance
\begin{equation}
    \label{eq:standard_gp_posterior}
    \begin{split}
        \mathbb{E}[\textbf{f}^{*} | X^{*}, X, Y] &= \mu(X^{*}) + K(X^{*}, X)[K(X, X) + \sigma^{2}I_{N}]^{-1}(Y - \mu(X)), \\
        \text{cov}(\textbf{f}^{*} | X^{*}, X, Y) & = K(X^{*}, X^{*}) - K(X^{*}, X)[K(X, X) + \sigma^{2}I_{N}]^{-1}K(X, X^{*}).
    \end{split}
\end{equation}

Here $\textbf{f}^{*} | X^{*}$ is the predictive distribution of $f(X^{*})$ and 
$\textbf{f}^{*} | X^{*}, X, Y$ is the predictive posterior given the data $X, Y$.
For further details see e.g. \cite{Rasmussen:2005:GPML}.

\subsection{Linear Operations on Gaussian Processes}
\label{sec:linop}
Let $\linop{}$ be a linear operator on realizations of $f \sim \mathcal{GP}(\mu(\x), K(\x, \x'))$. As GPs are closed under linear 
operators \citep{Rasmussen:2005:GPML, papoulis:2002:rvsp}, $\linop{}f$ is still a GP
\footnote{We assume here that $\linop{}f$ exists. For instance, if $\linop{}$ involves differentiation then 
the process $f$ must be differentiable. See e.g. \citep{Adler:1981:geometry} for details on proving existence.}. 
We will assume that the operator produces functions with range in $\mathbb{R}^{n_{c}}$, but where the input domain $\mathbb{R}^{n_{x}}$
is unchanged. That is, the operator produces functions from $\mathbb{R}^{n_{x}}$ to $\mathbb{R}^{n_{c}}$. 
This type of operators on GPs has also been considered by \citet{Sarkka:2011:SPDE} with applications to stochastic partial differential equations. 
The mean and covariance of $\linop{}f$ are given by applying $\linop{}$ to the mean and covariance of the argument:
\begin{equation}
    \label{eq:linop_mean}
    \begin{split}
        \mathbb{E}[\linop{}f(\x)] &= \linop{}\mu(\x) : \mathbb{R}^{n_{x}} \rightarrow \mathbb{R}^{n_{c}}, \\
        \text{cov}(\linop{}f(\x), \linop{}f(\x')) & = \linop{} K(\x, \x') \linop{}^{T} : \mathbb{R}^{n_{x} \times n_{x}} \rightarrow \mathbb{R}^{n_{c} \times n_{c}}, 
    \end{split}
\end{equation}
and the cross-covariance is given as 
\begin{equation}
    \label{eq:linop_cov}
    \begin{split}
        \text{cov}(\linop{}f(\x), f(\x')) & = \linop{} K(\x, \x') :  \mathbb{R}^{n_{x} \times n_{x}} \rightarrow \mathbb{R}^{n_{c}},\\
        \text{cov}(f(\x), \linop{}f(\x')) & = K(\x, \x') \linop{}^{T}:  \mathbb{R}^{n_{x} \times n_{x}} \rightarrow \mathbb{R}^{n_{c}}. 
    \end{split}
\end{equation}
The notation $\linop{} K(\x, \x')$ and $K(\x, \x') \linop{}^{T}$ is used to indicate when the operator acts on  
$K(\x, \x')$ as a function of $\x$ and $\x'$ respectively. That is, $\linop{} K(\x, \x') = \linop{} K(\x, \cdot)$ and $K(\x, \x') \linop{} = \linop{} K(\cdot, \x')$.
With the transpose operator the latter becomes $K(\x, \x') \linop{}^{T} = (\linop{} K(\cdot, \x'))^{T}$.
In the following sections we make use of the predictive distribution \eqref{eq:standard_gp_posterior},
where observations correspond to the transformed GP under $\linop{}$.

\section{Gaussian Processes with Linear Inequality Constraints}
\label{sec:CGP}
Following Section \ref{sec:GPR} and Section \ref{sec:linop}, we let $f \sim \mathcal{GP}(\mu(\x), K(\x, \x'))$ be a GP over real valued functions 
on $\mathbb{R}^{n_{x}}$, and $\linop{}$ a linear operator producing functions from $\mathbb{R}^{n_{x}}$ to $\mathbb{R}^{n_{c}}$. 
The matrix $X$ and the vector $Y$ will represent $N$ noise perturbed observations: $y_{i} = f(\x_{i}) + \varepsilon_{i}$ 
with $\varepsilon_{i}$ i.i.d. $\mathcal{N}(0, \sigma^{2})$ for $i = 1, \dots, N$. 

We would like to model the posterior GP conditioned on the observations $X, Y$, and on the event that $a(\x) \leq \linop{}f(\x) \leq b(\x)$
for two functions $a(\x), b(\x) : \mathbb{R}^{n_{x}} \rightarrow (\mathbb{R}\cup \{ -\infty, \infty \})^{n_{c}}$, where $a_{i}(\x) < b_{i}(\x)$
for all $\x \in \mathbb{R}^{n_{x}}$ and $i = 1, \dots, n_{c}$. 
To achieve this approximately, we start by assuming that the constraint $a(\x) \leq \linop{}f(\x) \leq b(\x)$ only holds at 
a finite set of inputs $\x^{v}_{1}, \dots, \x^{v}_{\NVI}$ that we refer to as \emph{virtual observation locations}. 
Later, we will consider how to specify the set of virtual observation locations such that the constraint 
holds for any $\x$ with sufficiently high probability. 
Furthermore, we will also assume that \emph{virtual observations} of the transformed process, $\linop{}f(\x^{v}_{i})$, comes with 
additive white noise with variance $\sigma^{2}_{v}$. 
We can write this as $a(X^{v}) \leq \linop{}f(X^{v}) + \varepsilon^{v} \leq b(X^{v})$,
where $X^{v} = [\x^{v}_{1}, \dots, \x^{v}_{\NVI}]^{T}$ is the matrix containing the virtual observation locations
and $\varepsilon^{v}$ is a multivariate Gaussian with diagonal covariance of elements $\sigma^{2}_{v}$. 

We will make use of the following notation: 
Let $\widetilde{C}(X^{v}) \in \mathbb{R}^{\NVI \times n_{c}}$ be the matrix with rows $(\widetilde{C}(X^{v}))_{i} = \linop{}f(\x^{v}_{i})
+ \varepsilon^{v}_{i}$ for i.i.d. $\varepsilon^{v}_{i} \sim \mathcal{N}(\textbf{0}, \sigma_{v}^{2} I_{n_{c}})$,  
and let $C(X^{v})$ denote the event $C(X^{v}) := \cap_{i=1}^{\NVI} \{ a(\x^{v}_{i}) \leq (\widetilde{C}(X^{v}))_{i} \leq b(\x^{v}_{i}) \}$.
$C(X^{v})$ thus represents the event that the constraint $a(\x) \leq \linop{}f(\x) + \varepsilon^{v} \leq b(\x)$ is satisfied for all points in $X^{v}$,
and it is defined through the latent variable $\widetilde{C}(X^{v})$. 

In summary, the process we will consider is stated as
\begin{equation*}
    f | X, Y, X^{v}, C(X^{v}) := f | f(X) + \varepsilon = Y, a(X^{v}) \leq \linop{}f(X^{v}) + \varepsilon^{v} \leq b(X^{v}), 
\end{equation*}
where $f$ is a Gaussian process, $X, Y$ is the training data and $X^{v}$ are the locations where the 
transformed process $\linop{}f + \varepsilon^{v}$ is bounded. The additive noise $\varepsilon$ and $\varepsilon^{v}$ are multivariate
Gaussian with diagonal covariance matrices of elements $\sigma^{2}$ and $\sigma_{v}^{2}$ respectively.

Here we assume that observations of all parts of $\linop{}f$ comes with i.i.d. white noise with variance $\sigma_{v}^{2}$. 
The reason for this is mainly for numerical stability, where we in computations will choose a tiny variance to approximate noiseless observations. 
Similarly, $\sigma^{2}$ may be chosen as a fixed small number for interpolation in the standard GP regression setting. 
In the following derivations, the results for exact noiseless observations can be obtained by setting the relevant variance to zero. 

We also assume that any sub-operator of $\linop{}$ is constrained at the 
same set of virtual locations $X^{v}$. 
This is mainly for notational convenience, 
and this assumption will be relaxed in Section \ref{sec:separate_obs}. In the following, we 
let $N_{v}$ denote the total number of virtual observation locations. Here 
$N_{v} = \NVI \cdot n_{c}$ for now, whereas we will later consider $N_{v} = \sum_{i = 1}^{n_{c}} \NVI_{i}$ 
where the i-th sub-operator is associated with $\NVI_{i}$ virtual observation locations. 

\subsection{Posterior Predictive Distribution}
\label{sec:CGP_posterior}
Our goal is to obtain the posterior predictive distribution $\f^{*} | X^{*}, X, Y, X^{v}, C(X^{v})$. 
That is: the distribution of $\f^{*} = f(X^{*})$ for some new inputs $X^{*}$, conditioned on the observed data 
$Y = f(X) + \varepsilon$ and the constraint 
$a(X^{v}) \leq \linop{}f(X^{v}) + \varepsilon^{v} \leq b(X^{v})$.

To simplify the notation we write $\f^{*} |Y, C$,
excluding the dependency on inputs $X, X^{*}$ and $X^{v}$ 
(as well as any hyperparameter of the mean and covariance function). 
The posterior predictive distribution is given by marginalizing over the latent variable $\widetilde{C}$:
\begin{equation*}
    \begin{aligned}
        &p(\f^{*}, C | Y ) = p(\f^{*} | C, Y)p(C | Y), \\
        &p(\f^{*} | C, Y) = \int_{a(X^{v})}^{b(X^{v})} p(\f^{*} | \widetilde{C}, Y) p(\widetilde{C} | Y) d \widetilde{C}, \\  
        &p(C | Y) = \int_{a(X^{v})}^{b(X^{v})} p(\widetilde{C} | Y) d \widetilde{C},
    \end{aligned}
\end{equation*}
where the limits correspond to the hyper-rectangle in $\mathbb{R}^{N_{v}}$ given by 
the functions $a(\cdot)$ and $b(\cdot)$ evaluated at each $\x^{v} \in X^{v}$. The predictive distribution and the probability $p(C | Y)$ are given 
in Lemma \ref{lemma:posterior}. $p(C | Y)$ is of interest, as it is 
the probability that the constraint holds at $X^{v}$ given the data $Y$. 
\\

In the remainder of the paper we will use the shortened 
notation $\mu^{*} = \mu(X^{*})$, $\mu = \mu(X)$, 
$\mu^{v} = \mu(X^{v})$ 
and $K_{X, X'} = K(X, X')$.
For vectors with elements in $\mathbb{R}^{n_{c}}$, such as $\linop{}\mu^{v}$, we interpret this elementwise.
E.g. $\linop{}\mu^{v}(X^{v})$ is given by the column vector $[\linop{}\mu(\x^{v}_{1})_{1},\allowbreak \dots,\allowbreak \linop{}\mu(\x^{v}_{1})_{n_{c}},\allowbreak 
\dots,\allowbreak \linop{}\mu(\x^{v}_{\NVI})_{1},\allowbreak \dots,\allowbreak \linop{}\mu(\x^{v}_{\NVI})_{n_{c}}]$.

We start by deriving the posterior predictive distribution $\textbf{\textup{f}}^{*}$ at some new locations 
$X^{*}$. The predictive distribution is represented by a Gaussian, $\textbf{\textup{f}}^{*} | Y, C \sim \mathcal{N}(\mu(\textbf{C}), \Sigma)$,
for some fixed covariance matrix $\Sigma$ and a mean $\mu(\textbf{C})$ that depends on the random variable 
$\textbf{C} = \widetilde{C} | Y, C$. The variable $\widetilde{C} =  \linop{}f(X^{v}) + \varepsilon^{v}$ remains 
Gaussian after conditioning on the observations $Y$, i.e. $\widetilde{C} | Y \sim \mathcal{N}(\nu_{c}, \Sigma_{c})$ with some 
expectation $\nu_{c}$ and covariance matrix $\Sigma_{c}$ that can be computed using (\ref{eq:linop_mean}, \ref{eq:linop_cov}).
Applying the constraints represented by the event $C$ on the random variable $\widetilde{C} | Y$ just means
restricting $\widetilde{C} | Y$ to lie in the hyper-rectangle defined by the bounds $a(X^{v})$ and $b(X^{v})$.
This means that $\textbf{C} = \widetilde{C} | Y, C$ is a truncated multivariate Gaussian, 
$\textbf{C} \sim \mathcal{TN}( \nu_{c} , \Sigma_{c},a(X^{v}), b(X^{v}))$.  
The full derivation of the distribution parameters of $\textbf{C}$ and $\textbf{\textup{f}}^{*} | Y, C$ are 
given in Lemma \ref{lemma:posterior} below, whereas Lemma \ref{lemma:posterior_chol} provides an alternative algorithmic 
representation suitable for numerical implementation. 

\begin{lemma}
    \label{lemma:posterior}
    The predictive distribution $\textbf{\textup{f}}^{*} | Y, C$ is a compound Gaussian 
    with truncated Gaussian mean: 
    \begin{equation}
        \label{eq:posterior_1}
        \textbf{\textup{f}}^{*} | Y, C \sim \mathcal{N}(\mu^{*} + A(\textbf{C} - \linop{} \mu^{v}) + B(Y - \mu), \Sigma ),
    \end{equation}
    \begin{equation}
        \label{eq:posterior_2}
        \textbf{C} = \widetilde{C} | Y, C \sim \mathcal{TN}( \linop{} \mu^{v} + A_{1}(Y - \mu), B_{1}, a(X^{v}), b(X^{v}) ),
    \end{equation}
    where $\mathcal{TN}(\cdot, \cdot, a, b)$ is the Gaussian $\mathcal{N}(\cdot, \cdot)$ conditioned on the hyper-rectangle 
    $[a_{1}, b_{1}] \times \cdots \times [a_{k}, b_{k}]$, and
    \begin{gather*}
        \begin{array}{ll}
            A_{1} = (\linop{} K_{X^{v}, X})(K_{X, X} + \sigma^{2}I_{N})^{-1}, & B_{1} = \linop{} K_{X^{v}, X^{v}} \linop{}^{T} + \sigma^{2}_{v} I_{N_{v}} - A_{1} K_{X, X^{v}}\linop{}^{T}, \\
            A_{2} =  K_{X^{*}, X}(K_{X, X} + \sigma^{2}I_{N})^{-1}, & B_{2} = K_{X^{*}, X^{*}} - A_{2} K_{X, X^{*}}, \\ 
            & B_{3} = K_{X^{*}, X^{v}}\linop{}^{T} - A_{2} K_{X, X^{v}} \linop{}^{T}, \\ 
        \end{array} \\
        \begin{array}{lll}
            A = B_{3}B_{1}^{-1}, & B = A_{2} - AA_{1}, & \Sigma = B_{2} - AB_{3}^{T}.
        \end{array}
    \end{gather*}   
    Moreover, the probability that the unconstrained version of $\textbf{C}$
    falls within the constraint region, $p(C | Y)$, is given by
    \begin{equation}
        \label{eq:constrprob}
        p(C | Y) = p \left( a(X^{v}) \leq \mathcal{N}( \linop{} \mu^{v} + A_{1}(Y - \mu), B_{1})  \leq b(X^{v}) \right), 
    \end{equation}
    and the unconstrained predictive distribution is 
    \begin{equation*}
        \textbf{\textup{f}}^{*} | Y \sim \mathcal{N}(\mu^{*} + A_{2}(Y - \mu), B_{2} ).
    \end{equation*}

\end{lemma}

The derivation in Lemma \ref{lemma:posterior} is based on conditioning the 
multivariate Gaussian $(\f^{*}, Y, \widetilde{C})$, and the proof is given in Appendix \ref{app:proof:posterior}.
For practical implementation the matrix inversions involved in Lemma \ref{lemma:posterior} 
may be prone to numerical instability. A numerically stable alternative is given in Lemma \ref{lemma:posterior_chol} below.

In the following lemma, \textit{Chol}$(K)$ is the lower triangular Cholesky 
factor of a matrix $K$. We also let $R = (P \setminus Q)$ denote the solution to the linear 
system $PR = Q$ for matrices $P$ and $Q$, which may be efficiently computed when $P$ is triangular
using forward or backward substitution. 

\begin{lemma}
    \label{lemma:posterior_chol}
    Let 
    $L = Chol(K_{X, X} + \sigma^{2}I_{N})$, 
    $v_{1} = L \setminus  K_{X, X^{v}} \linop{}^{T}$ and
    $v_{2} = L \setminus K_{X, X^{*}}$.
    \\

    Then the matrices in Lemma \ref{lemma:posterior} can be computed as
    \begin{equation*}
        \begin{array}{ll}
            A_{1}  = (L^{T} \setminus v_{1})^{T}, & B_{1} = \linop{}  K_{X^{v}, X^{v}} \linop{}^{T} + \sigma^{2}_{v} I_{N_{v}} - v_{1}^{T}v_{1}, \\
            A_{2}  = (L^{T} \setminus v_{2})^{T}, & B_{2} = K_{X^{*}, X^{*}} - v_{2}^{T}v_{2}, \\ 
            & B_{3} = K_{X^{*}, X^{v}} \linop{}^{T} - v_{2}^{T}v_{1}.
        \end{array}
    \end{equation*}
    Moreover, $B_{1}$ is symmetric and positive definite. By letting $L_{1} = Chol(B_{1})$ and $v_{3} = L_{1} \setminus B_{3}^{T}$
    we also have
    \begin{equation*}
        \begin{array}{lll}
            A = (L_{1}^{T} \setminus v_{3})^{T}, &  B = A_{2} - AA_{1}, & \Sigma = B_{2} - v_{3}^{T}v_{3}.
        \end{array}
    \end{equation*}

\end{lemma}

The proof is given in Appendix \ref{app:proof:chol}. The numerical complexity of the procedures in Lemma \ref{lemma:posterior_chol} is $n^{3} / 6$ for Cholesky factorization 
of $n \times n$ matrices 
and $mn^{2} / 2$ for solving triangular systems where the unknown matrix is $n \times m$. In the derivation of Lemma \ref{lemma:posterior} and Lemma \ref{lemma:posterior_chol}, the order of operations was chosen such that the first Cholesky factor $L = Chol(K_{X, X} + \sigma^{2}I_{N})$ only depends on $X$. 
This is convenient in the case where the posterior $\textbf{\textup{f}}^{*} | Y, C$ is calculated multiple times
for different constraints $C$ or virtual observations $X^{v}$, but where the data $X, Y$ remain unchanged. 

\subsection{Sampling from the Posterior Distribution}
\label{sec:CGP_sampling}
In order to sample from the posterior we can first sample from the constraint distribution \eqref{eq:posterior_2}, 
and then use these samples in the mean of \eqref{eq:posterior_1} to create the final samples of $\f^{*} | Y, C$.

To generate $k$ samples of the posterior at $M$ new input locations, $[\x^{*}_{1}, \dots, \x^{*}_{M}]^{T} = X^{*}$, 
we use the following procedure

\begin{alg}
    \label{alg:sampling}
    Sampling from the posterior distribution

    \begin{enumerate}
        \item Find a matrix $Q$ s.t. $Q^{T}Q = \Sigma \in \mathbb{R}^{M \times M}$, e.g. by Cholesky or a spectral decomposition. 
        \item Generate $\widetilde{C}_{k}$, a $N_{v} \times k$ matrix where each column is a sample of $\widetilde{C} | Y, C$ 
        from the distribution in \eqref{eq:posterior_2}.
        \item Generate $U_{k}$, a $M \times k$ matrix with $k$ samples from the standard normal $\mathcal{N}(\textbf{0}, I_{M}$).
        \item The $M \times k$ matrix where each column in a sample from $\textnormal{\f}^{*} | Y, C$ is then obtained by
        \begin{equation*}
            [\mu^{*} + B(Y - \mu)] \oplus_{col} [A(- \linop{} \mu^{v} \oplus_{col} \widetilde{C}_{k}) + QU_{k}],
        \end{equation*} 
        where $\oplus_{col}$ means that the $M \times 1$ vector on the left hand side is 
        added to each column of the $M \times k$ matrix on the right hand side. 
    \end{enumerate}

\end{alg}

This procedure is based on the well-known method for sampling from multivariate Gaussian distributions, 
where we have used the property that in the distribution of $\f^{*} | Y, C$, only the mean depends on 
samples from the constraint distribution. 

The challenging part of this procedure is the second step where samples have to be drawn from 
a truncated multivariate Gaussian. The simplest approach is by rejection sampling, i.e. 
generating samples from the normal distribution and rejection those that fall outside the 
bounds. In order to generate $m$ samples with rejection sampling, the expected number 
of samples needed is $m / p(C | Y)$, where the acceptance rate is the probability $p(C | Y)$ 
given in \eqref{eq:constrprob}. If the acceptance rate is low, then rejection sampling 
becomes inefficient, and an alternative approach such as Gibbs sampling \citep{Kotecha:1999:Gibbs} is typically used.
In our numerical experiments (presented in Section \ref{sec:num_exp}) we made use of a new method 
based on simulation via minimax tilting by \citet{Botev:2017:minimax_tilting}, developed for 
high-dimensional exact sampling. \citet{Botev:2017:minimax_tilting} prove 
strong efficiency properties and demonstrate accurate simulation in dimensions $d \sim 100$ with small 
acceptance probabilities ($\sim 10^{-100}$), that take about the same time as one cycle of Gibbs sampling.    
For higher dimensions in the thousands, the method is used to accelerate existing Gibbs samplers by sampling jointly 
hundreds of highly correlated variables. 
In our experiments, we 
experienced that this method worked well in cases where Gibbs sampling was challenging. A detailed comparison 
with other sampling alternatives for an application similar to ours is also given in \citep{LpezLopera:2017:CGP}.   
An important observation in Algorithm \ref{alg:sampling} is that for inference at a new set of input locations $X^{*}$, when 
the data $X, Y$ and virtual observation locations $X^{v}$ are unchanged, the samples generated in step \textit{2}
can be reused. 

\subsection{Parameter Estimation}
\label{sec:CGP_MLE}
To estimate the parameters of the CGP we make use of the marginal maximum likelihood approach (MLE). 
We define the marginal likelihood function of the CGP as
\begin{equation}
    \label{eq:lik_constr}
    L(\theta) = p(Y , C | \theta)  = p(Y | \theta) p(C | Y, \theta),
\end{equation}
i.e. as the probability of the data $Y$ and constraint $C$ combined, given the set of parameters represented by $\theta$.
We assume that both the mean and covariance function of the GP prior \eqref{eq:gp_prior} $\mu (\x | \theta)$ and $K(\x, \x' | \theta)$
may depend on $\theta$. 
The log-likelihood, $l(\theta) = \ln p(Y | \theta) + \ln p(C | Y, \theta)$, is thus given as the sum of the unconstrained log-likelihood, 
$\ln p(Y | \theta)$, which is optimized in unconstrained MLE, and $\ln p(C | Y, \theta)$, which is the probability that the 
constraint holds at $X^{v}$ given in \eqref{eq:constrprob}. 

In \citep{Bachoc:2018:GP_MLE} the authors study
the asymptotic distribution of the MLE for shape-constrained GPs, and show that for large sample sizes the 
effect of including the constraint in the MLE is negligible. But for small or moderate sample sizes the constrained MLE
is generally more accurate, so taking the constraint into account is beneficial. 
However, due to the added numerical complexity in optimizing a function that includes the term $\ln p(C | Y, \theta)$,
it might not be worthwhile. Efficient parameter estimation using the full likelihood \eqref{eq:lik_constr} 
is a topic of future research.
In the numerical experiments presented in this paper, we therefore make use of the unconstrained MLE. 
This also makes it possible to compare models with and without constraints in a more straightforward manner. 

\subsection{Finding the Virtual Observation Locations}
\label{sec:find_xv}
For the constraint to be satisfied locally at any input location 
in some bounded set $\Omega \subset \mathbb{R}^{n_{x}}$ with sufficiently high probability, 
the set of virtual observation locations $X^{v}$ has to be
sufficiently dense. We will specify a target probability $p_{\text{target}} \in [0, 1)$ and find a set $X^{v}$, such that
when the constraint is satisfied at all virtual locations in $X^{v}$, the probability that the constraint is satisfied 
for any $\x$ in $\Omega$ is at least $p_{\text{target}}$. The number of virtual observation locations needed 
depends on the smoothness properties of the kernel, and for a given kernel it is of interest
to find a set $X_{v}$ that is effective in terms of numerical computation. 
As we need to sample from a truncated Gaussian involving cross-covariances between all elements 
in $X^{v}$, we would like the set $X^{v}$ to be small, and also 
to avoid points in $X^{v}$ close together that could lead to high serial correlation. 

Seeking an optimal set of virtual observation locations has also been discussed in 
\citep{WangBerger:2016:CGP, Golchi:2015:CGP, Riihimaki:2010:CGP, DaVeiga:2012:CGP, DaVeiga:2017:CGP}, 
and the intuitive idea is to iteratively place virtual observation locations where the probability 
that the constraint holds is low. The general approach presented in this section is most similar to 
that of \citet{WangBerger:2016:CGP}. In Section \ref{sec:separate_obs} we extend this to derive a 
more efficient method for multiple constraints. 

In order to estimate the probability that the constraint holds at some new location $\x^{*} \in \Omega$, 
we first derive the posterior distribution of the constraint process. 

\begin{lemma}
    \label{lemma:constr_posterior}
    The predictive distribution of the constraint $\linop{}f(\x^{*})$ 
    for some new input $\x^{*} \in \mathbb{R}^{n_{x}}$, condition on the data $Y$ 
    is given by 
    \begin{equation}
        \label{eq:constr_posterior_data}
        \linop{}f(\x^{*}) | Y \sim \mathcal{N}(\linop{}\mu^{*} + \widetilde{A}_{2}(Y - \mu), \widetilde{B}_{2}), 
    \end{equation}
    and when $\linop{}f(\x^{*})$ is conditioned on both the data and virtual constraint observations, $X, Y$ and $X^{v}, C(X^{v})$,
    the posterior becomes 
    \begin{equation}
        \label{eq:constr_posterior}
        \linop{}f(\x^{*}) | Y, C \sim \mathcal{N}(\linop{}\mu^{*} + \widetilde{A}(\textbf{C} - \linop{} \mu^{v}) + \widetilde{B}(Y - \mu), \widetilde{\Sigma} ).
    \end{equation}
    Here $L$, $v_{1}$, $A_{1}$, $B_{1}$ and $L_{1}$ are defined as in Lemma \ref{lemma:posterior_chol}
    , $\textbf{C}$ is the distribution in \eqref{eq:posterior_2} and
    \begin{gather*}
        \begin{array}{ll}
            \widetilde{v}_{2} = L \setminus  K_{X, \x^{*}} \linop{}^{T}, & \widetilde{B}_{2} = \linop{}K_{\x^{*}, \x^{*}} \linop{}^{T} - \widetilde{v}_{2}^{T}\widetilde{v}_{2}, \\
            \widetilde{A}_{2} = (L^{T} \setminus \widetilde{v}_{2})^{T}, & \widetilde{B}_{3} = \linop{}K_{\x^{*}, X^{v}} \linop{}^{T}- \widetilde{v}_{2}^{T}v_{1},\\
            & \widetilde{v}_{3} = L_{1} \setminus \widetilde{B}_{3}^{T},
        \end{array} \\
        \begin{array}{lll}
            \widetilde{A} = (L_{1}^{T} \setminus \widetilde{v}_{3})^{T}, & \widetilde{B} = \widetilde{A}_{2} - \widetilde{A}A_{1}, & \widetilde{\Sigma} = \widetilde{B}_{2} - \widetilde{v}_{3}^{T}\widetilde{v}_{3}.
        \end{array}
    \end{gather*}
\end{lemma}

The proof is given in Appendix \ref{app:proof:constr_posterior}. 
The predictive distribution in Lemma \ref{lemma:constr_posterior} was defined for a single input $\x^{*} \in \mathbb{R}^{n_{x}}$, and 
we will make use of the result in this context. But we could just as well consider an input matrix
$X^{*}$ with rows $\x^{*}_{1}, \x^{*}_{2}, \dots$, where the only change in Lemma \ref{lemma:constr_posterior} is 
to replace $\x^{*}$ with $X^{*}$. In this case we also note that the variances, $\text{diag}( \widetilde{\Sigma})$, 
is more efficiently computed as $\text{diag}( \widetilde{\Sigma}) =  \text{diag}(\linop{}K_{X^{*}, X^{*}}\linop{}^{T})
- \text{diag}(\widetilde{v}_{2}^{T}\widetilde{v}_{2}) - \text{diag}(\widetilde{v}_{3}^{T}\widetilde{v}_{3})$ where 
we recall that $\text{diag}(v^{T}v)_{i} = \sum_{j} v_{i, j}^{2}$ for $v^{T} = [v_{i, j}]$.

Using the posterior distribution of $\linop{}f$ in Lemma \ref{lemma:constr_posterior} we define the constraint probability
$p_{c} : \mathbb{R}^{n_{x}} \rightarrow [0, 1]$ as 
\begin{equation}
    \label{eq:pc}
    p_{c}(\x) = P \left( a(\x) - \nu < \xi(\x, X^{v}) < b(\x) +\nu \right),
\end{equation}
where $\xi(\x, X^{v}) = \linop{}f(\x^{*}) | Y$ for $X^{v} = \emptyset$
and $\xi(\x, X^{v}) = \linop{}f(\x^{*}) | Y, C$ otherwise. The quantity $\nu$ is a non-negative 
fixed number that is included to ensure that it will be possible to increase $p_{c}$ using 
observations with additive noise. When we use virtual observations $\widetilde{C}(\x) = \linop{}f(\x^{*}) + \varepsilon^{v}$
that come with noise $\varepsilon^{v} \sim \mathcal{N}(0, \sigma_{v}^{2})$, we can use
$\nu = \text{max} \{ \sigma_{v}\Phi^{-1}(p_{\text{target}}), 0 \}$ where $\Phi(\cdot)$ is the normal cumulative distribution function.
Note that $\sigma_{v}$, and in this case $\nu$, will be small numbers included mainly for numerical stability.
In the numerical examples presented in this paper this noise variance was set to $10^{-6}$.

In the case where $X^{v} = \emptyset$, computation of \eqref{eq:pc} is straightforward 
as $\xi(\x, X^{v})$ is Gaussian. Otherwise, we will rely on the following estimate of $p_{c}(\x)$: 
\begin{equation}
    \label{eq:p_hat}
    \hat{p}_{c}(\x) = \frac{1}{m} \sum_{j = 1}^{m} P \left( a(\x) - \nu < (\linop{}f(\x) | Y, C_{j}) < b(\x) + \nu \right),
\end{equation}
where $C_{1}, \dots, C_{m}$ are $m$ samples of $\textbf{C}$ given in \eqref{eq:posterior_2}.

We outline an algorithm for finding a set of virtual observation locations $X^{v}$, such that the probability that the constraint holds locally
at any $\x \in \Omega$ is at least $p_{\text{target}}$ for some specified
set $\Omega \subset \mathbb{R}^{n_{x}}$ and $p_{\text{target}} \in [0, 1)$. That is, $\min_{\x \in \Omega} p_{c}(\x) \geq p_{\text{target}}$.
The algorithm can be used starting with no initial virtual observation locations, $X^{v} =  \emptyset$,
or using some pre-defined set $X^{v} \neq \emptyset$. The latter may be useful e.g. if the 
data $X, Y$ is updated, in which case only a few additions to the previous set $X^{v}$ might be needed. 

\begin{alg}
    \label{alg:find_xv}
    Finding locations of virtual observations $X^{v}$ s.t. $\hat{p}_{c}(\x) \geq p_{\text{target}}$ for all $\x \in \Omega$.

    \begin{enumerate}
        \item Compute $L = Chol(K_{X, X} + \sigma^{2}I_{N})$.
        \item Until convergence do:
        \begin{enumerate}
            \item If $X^{v} \neq \emptyset$ compute $A_{1}$ and $B_{1}$ as defined in Lemma \ref{lemma:posterior_chol},
            and generate $m$ samples $C_{1}, \dots, C_{m}$ of $\textbf{C}$ given in \eqref{eq:posterior_2}.
            \item If $X^{v} = \emptyset$ compute $(\x^{*}, p^{*}) = (\argmin p_{c}(\x), p_{c}(\x^{*}))$. Otherwise compute  
            $(\x^{*}, p^{*}) = (\argmin \hat{p}_{c}(\x), \hat{p}_{c}(\x^{*}))$ with $\hat{p}_{c}$ defined as in 
            \eqref{eq:p_hat}, using the samples generated in step (a).
            \item Terminate if $p^{*} \geq p_{\text{target}}$, 
            otherwise update $X^{v} \rightarrow X^{v} \cup \{ \x^{*} \}$.
        \end{enumerate}
    \end{enumerate}
\end{alg}

The rate of convergence of Algorithm \ref{alg:find_xv} relies on the probability that the constraint holds initially, 
$P \left( a(\x) < (\linop{}f(\x) | Y) < b(\x) \right)$, and for practical application one may monitor
$p^{*}$ as a function of the number of virtual observation locations, $|X^{v}|$, to find an appropriate stopping criterion. 

With the exception of low dimensional input $\x$, the optimization step $\x^{*} = \argmin \hat{p}_{c}(\x)$ is 
in general a hard non-convex optimization problem. 
But with respect to how $\x^{*}$ and $ p^{*}$ are used in the algorithm, some 
simplifications can be justified. 
First, we note that when computing $\hat{p}_{c}(\x)$ with \eqref{eq:p_hat} for multiple $\x = \x_{1}, \x_{2}, \dots$, 
the samples $C_{1}, \dots, C_{m}$ are reused. 
It is also not necessary to find the the absolute minimum, as long as a \emph{small enough} value is found in each iteration. 
Within the global optimization one might therefore decide to stop after the first occurrence of $\hat{p}_{c}(\x)$ less 
than some threshold value. With this idea one could also search over finite candidate 
sets $\Omega \subset \mathbb{R}^{n_{x}}$, using a fixed number of random points in $\mathbb{R}^{n_{x}}$.
This approach might produce a larger set $X^{v}$, but where the selection of $\x^{*}$ is 
faster in each iteration. 
Some of the alternative strategies for locating $\x^{*}$ in Algorithm \ref{alg:find_xv} 
are studied further in our numerical experiments in Section \ref{sec:num_exp}.

With the above algorithm we aim to impose constraints on some bounded set $\Omega \subset \mathbb{R}^{n_{x}}$. 
Here $\Omega$ has to be chosen with respect to both training and test data. 
For a single boundedness constraint, it might be sufficient that the constraint only holds at the points $\x \in \mathbb{R}^{n_{x}}$
that will be used for prediction. But if we consider constraints related to monotonicity (see Example 1, Section \ref{sec:num_exp}), 
dependency with respect to the latent function's properties at the training locations is lost with this strategy.
In the examples we give in this paper we consider a convex set $\Omega$, in particular $\Omega = [0, 1]^{n_{x}}$, 
and assume that training data, test data and any input relevant for prediction lies within $\Omega$. 

\subsection{Separating Virtual Observation Locations for Sub-operators}
\label{sec:separate_obs}
Let $\linop{}$ be a linear operator defined by the column vector $[\sublinop{1}, \dots, \sublinop{k}]$,
where each $\sublinop{i}$ is a linear operator leaving both the domain and range of its argument unchanged, 
i.e. $\sublinop{i}$ produces functions from $\mathbb{R}^{n_{x}}$ to $\mathbb{R}$, subjected to an interval constraint $[a_{i}(\x), b_{i}(\x)]$. 
Until now we have assumed that the constrain holds at a set of virtual observation locations $X^{v}$, which means that 
$a_{i}(X^{v}) \leq \sublinop{i}f(X^{v}) \leq b_{i}(X^{v})$ for all $i = 1, \dots, k$. 

However, it might not be necessary to constrain each of the sub-operators $\sublinop{i}$ at the same points $\x^{v} \in X^{v}$. 
Intuitively, constraints with respect to $\sublinop{i}$ need only be imposed at locations where $p(\sublinop{i}f(\x) \notin [a_{i}(\x), b_{i}(\x)])$
is large. To accommodate this we let $X^{v}$ be the concatenation of the matrices $X^{v, 1}, \dots, X^{v, k}$
and define $\linop{}^{T}f(X^{v}) = [\sublinop{1}^{T}f(X^{v, 1}), \dots, \sublinop{k}^{T}f(X^{v, 1})]^{T}$.
This is equivalent to removing 
some of the rows in $\linop{}(\cdot)(X^{v})$, and all of the results in this paper still apply.
In this setting we can improve the algorithm in Section \ref{sec:find_xv} for finding the set of virtual observation locations 
by considering each sub-operator individually. 
This is achieved using the estimated partial constraint probabilities, $p_{c, i}(\x)$, that we defined as 
in \eqref{eq:pc} by considering only the i-th sub-operator. We may then use the estimate
\begin{equation}
    \label{eq:p_hat_i}
    \hat{p}_{c, i}(\x) = \frac{1}{m} \sum_{j = 1}^{m} P \left( a_{i}(\x) - \nu < (\linop{}f(\x) | Y, C_{j})_{i} < b_{i}(\x) + \nu \right),
\end{equation}
where $(\linop{}f(\x) | Y, C_{j})_{i}$ is the univariate Normal distribution given by the $i$-th row of $(\linop{}f(\x) | Y, C_{j})$,
and $C_{1}, \dots, C_{m}$ are $m$ samples of $\textbf{C}$ given in \eqref{eq:posterior_2} as before. 
Algorithm \ref{alg:find_xv} can then be improved by minimizing \eqref{eq:p_hat_i} with respect to 
both $\x$ and $i = 1, \dots k$. The details are presented in Appendix \ref{app:search_xv_i}, Algorithm \ref{alg:find_xvi}.

\subsection{Prediction using the Posterior Distribution}
For the unconstrained GP in this paper where the likelihood is given by Gaussian white noise, 
the posterior mean and covariance is sufficient to describe predictions as the posterior remains Gaussian.
It is also known that in this case there is a correspondence between the posterior mean of the GP
and the optimal estimator in the Reproducing Kernel Hilbert Space (RKHS) associated with the GP \citep{kimeldorf:1970:correspondence}. 
This is a Hilbert space of functions defined by the positive semidefinite kernel of the GP. Interestingly, a similar correspondence holds for the constrained case. \citet{Bay:2017:GeneralizationKW} 
show that for constrained interpolation, the Maximum \textit{A Posteriori} (MAP) or mode of the posterior 
is the optimal constrained interpolation function in the RKHS, 
and also illustrate in simulations that the unconstrained mean and constrained MAP coincide only when the unconstrained mean satisfies the constraint. 
This holds when the GP is constrained 
to a convex set of functions, which is the case in this paper where we condition on 
linear transformations of a function restricted to a convex set.

\subsection{An Alternative Approach based on Conditional Expectations}
\label{sec:moment_approach}
\citet{DaVeiga:2012:CGP, DaVeiga:2017:CGP} propose an approach for approximating the first two moments of the 
constrained posterior, $\f^{*} | Y, C$, using conditional expectations of the truncated multivariate Gaussian. 
This means, in the context of this paper, that the first two moments of $\f^{*} | Y, C$ are computed using 
the first two moments of the latent variable $\textbf{C}$. To apply this idea using the formulation of this paper, 
we can make use of the following result. 

\begin{corollary}
    \label{corr:moments}
    Let the matrices $A$, $B$, $\Sigma$ and the truncated Gaussian 
    random variable $\textbf{C}$ be as defined in Lemma \ref{lemma:posterior},
    and let $\nu, \Gamma$ be the expectation and covariance of $\textbf{C}$.
    Then the expectation and covariance of the predictive distribution $\f^{*} | Y, C$
    are given as 
    \begin{equation}
        \label{eq:post_mom}
        \begin{gathered}        
            \mathbb{E}(\f^{*} | Y, C) = \mu^{*} + A(\nu  - \linop{}\mu^{v}) + B(Y - \mu), \\
            \text{cov}(\f^{*} | Y, C) = \Sigma + A\Gamma A^{T}.
        \end{gathered}
    \end{equation}
    Moreover, if $\widetilde{A}$, $\widetilde{B}$ and $\widetilde{\Sigma}$ are the matrices defined in Lemma \ref{lemma:constr_posterior},
    then the expectation and variance of the predictive distribution of the constraint $\linop{}f(\x^{*}) | Y, C$ are given as
    \begin{equation}
        \label{eq:post_constr_mom}
        \begin{gathered}        
            \mathbb{E}(\linop{}f(\x^{*}) | Y, C) = \linop{}\mu^{*}  + \widetilde{A}(\nu  - \linop{}\mu^{v}) + \widetilde{B}(Y - \mu), \\
            \text{var}(\linop{}f(\x^{*}) | Y, C) = \widetilde{\Sigma} + \widetilde{A}\Gamma \widetilde{A}^{T}.
        \end{gathered}
    \end{equation}
\end{corollary}

The results follows directly from the distributions derived in Lemmas \ref{lemma:posterior} and \ref{lemma:constr_posterior}, and 
moments of compound distributions. A proof is included in Appendix \ref{app:proof:corr:moments} for completeness. 

\citet{DaVeiga:2012:CGP, DaVeiga:2017:CGP} make use of a Genz approximation 
\citep{Genz:1992:numerical_comp, Genz:1997:comparisson} to compute $\nu, \Gamma$ for inference using \eqref{eq:post_mom}.
They also introduce a crude but faster correlation-free approximation that can be used in the search for 
virtual observation locations. With this approach, \eqref{eq:post_constr_mom} is used where $\nu, \Gamma$ are 
computed under the assumption that $\text{cov}(\widetilde{C}|Y)$ is diagonal. We can state this 
approximation as follows: 
\begin{equation*}
            \nu_{i} \approx m_{i} + s_{i} \frac{\phi(\widetilde{a_{i}}) - \phi(\widetilde{b_{i}})}{\Phi(\widetilde{b_{i}}) - \Phi(\widetilde{a_{i}})}, \ \ \
            \Gamma_{i, i} \approx s_{i}^{2} \left[ 1 + \frac{\widetilde{a_{i}}\phi(\widetilde{a_{i}}) - \widetilde{b_{i}}\phi(\widetilde{b_{i}})}{\Phi(\widetilde{b_{i}}) - \Phi(\widetilde{a_{i}})}
            - \left( \frac{\phi(\widetilde{a_{i}}) - \phi(\widetilde{b_{i}})}{\Phi(\widetilde{b_{i}}) - \Phi(\widetilde{a_{i}})} \right)^{2} \right],        
\end{equation*}
where $m_{i}$ is the i-th component of $\mathbb{E}(\widetilde{C}|Y) = \linop{} \mu^{v} + A_{1}(Y - \mu)$, $s_{i} = \sqrt{\text{cov}(\widetilde{C}|Y)_{i, i}} = \sqrt{(B_{1})_{i, i}}$,
$\widetilde{a_{i}} = (a(X^{v})_{i} - m_{i}) / s_{i}$, $\widetilde{b_{i}} = (b(X^{v})_{i} - m_{i}) / s_{i}$,
$\phi$ and $\Phi$ are the pdf and cdf of the standard normal distribution and 
$\Gamma$ is diagonal with elements $\Gamma_{i, i}$. We will make use of these approximations in some of the examples in Section \ref{sec:num_exp}
for comparison. 

\subsection{Numerical Considerations}
\label{sec:numerical_cons}
For numerical implementation, we discuss some key considerations with the proposed model. 
One of the main issues with implementation of GP models in terms of numerical stability is 
related to covariance matrix inversion, which is why alternatives such as Cholesky factorization
are recommended in practice. This does however not alleviate problems related to ill-conditioned 
covariance matrices. This is a common problem in computer code emulation (zero observational noise) in particular, 
where training points might be 'too close to each other' in terms of the covariance function, leaving 
the covariance matrix close to degenerate as some of the observations become redundant. 
A common remedy is to introduce a 'nugget' term on the diagonal entries of the covariance matrix, 
in the form of additional white noise on the observations. This means using a small $\sigma > 0$ instead 
of $\sigma = 0$ in Equation \eqref{eq:standard_gp_posterior}, even when the observations are noiseless. In terms 
of matrix regularization this is equivalent to Tikhonov regularization. See for instance
\citet{Ranjan:2010:stable_GP_interp} and \citet{Andrianakis:2012:nugget_GP} which give a detailed 
discussion and recommendations for how to choose appropriate 
value for $\sigma$. In practice, a fixed small value is often used without further analysis,
as long as the resulting condition number is not too high. 
This approach can be justified since 
the use of a nugget term has a straightforward interpretation, as opposed to other alternatives such as pseudoinversion. 
In our experiments on noiseless regression we fix $\sigma^{2} = 10^{-6}$, 
as the error introduced by adding a variance of $10^{-6}$ to the observations is negligible.

Similarly, for the virtual observations used in this paper we make use of the noise parameter 
$\sigma_{v}$ to avoid ill-conditioning of the matrix $B_{1}$ defined in Lemma \ref{lemma:posterior}. 
$B_{1}$ is the covariance matrix of the transformed GP, $\widetilde{C}|Y$, and $B_{1}^{-1}$ together with  
$(K_{X, X} + \sigma^{2}I_{N})^{-1}$ are needed for all the posterior computations that involve constraints.
The virtual noise parameter $\sigma_{v}$ has a similar interpretation as $\sigma$, but where the 
artificial added noise acts on observations of the transformed process. Here $\sigma_{v} = 0$ means that 
the constraints are enforced with probability $1$, $\sigma_{v} > 0$ implies that the constraints 
are enforced in a soft way, and $\sigma_{v} \rightarrow \infty$ provides no constraint at all. 
In the numerical examples presented in this paper, a fixed value $\sigma_{v}^{2} = 10^{-6}$ has been used 
to approximate hard constraints with an error we find negligible. 

As for computational complexity, we may start by first looking at the operations involved in
computing the posterior predictive distribution 
at $M$ inputs $\x^{*}_{1}, \dots, \x^{*}_{M}$ (including covariances),
using Lemma \ref{lemma:posterior_chol}. 
We first make note of the operations needed in the unconstrained case, i.e. standard GP regression with Gaussian noise, for comparison.
If there are $N \geq M$ observations in the training set, then the complexity is dominated by the Cholesky factorization 
$L = Chol(K_{X, X} + \sigma^{2}I_{N})$, which require an order of $N^{3}$ operations and $N^{2}$ in memory. 
The Cholesky factor may be stored for subsequent predictions. Then, to compute the posterior predictive distribution 
at $M$ new inputs, the number of operations needed 
is dominated by matrix multiplication and solving triangular systems, of orders $NM^{2}$ and $N^{2}M$.
When a number $N_{v}$ of virtual observation locations are included, we are essentially dealing with the same computations as 
the standard GP regression, but with $N + N_{v}$ number of observations. I.e. the computations involved 
are of order $(N + N_{v})^{3}$ in time and $(N + N_{v})^{2}$ in memory. The order of operations in Lemma \ref{lemma:posterior_chol} 
was chosen such that the Cholesky factor $L$ that only depends on the training data can be reused. 
For a new set $X^{v}$ of size $N_{v}$, the computations needed for prediction at $M$ new locations $X^{*}$ will only 
require the Cholesky factorization $L_{1} = Chol(B_{1})$ of order $N_{v}^{3}$.
When both $L$ and $L_{1}$ are stored,
the remaining number of operations will be of order $N^{2}M$ or $N_{v}^{2}M$ for solving triangular systems, 
and $NM^{2}$, $N_{v}M^{2}$ or $NMN_{v}$ for matrix multiplications.

In order to sample from the posterior using Algorithm \ref{alg:sampling}, some additional steps are required. 
After the computations of Lemma \ref{lemma:posterior_chol} we continue to factorize the $M \times M$ covariance matrix $\Sigma$ 
and generate samples from the truncated Gaussian $\widetilde{C} | Y, C$. 
The complexity involved in sampling from this $N_{v}$-dimensional truncated Gaussian depends 
on the sampling method of choice, see Section \ref{sec:CGP_sampling}. We can combine $k$ of these samples with 
$k$ samples from a standard normal $\mathcal{N}(\textbf{0}, I_{M})$ to obtain samples of the final posterior, 
using an order of $MN_{v}k + M^{2}k$ operations. The total procedure of generating $k$ samples at $M \leq N$ new inputs 
is therefore dominated by matrix operations of order $(N + N_{v})^{3}$, $MN_{v}k$ and $ M^{2}k$, together 
with the complexity involved with sampling from a $N_{v}$-dimensional truncated Gaussian. 
For subsequent prediction it is convenient to here also reuse the samples generated 
from the truncated Gaussian, together with results that only involve $X$ and $X^{v}$. This means storing matrices
of size $N_{v} \times k$, $N \times N$ and $N_{v} \times N_{v}$. The remaining computations are then dominated 
by operations of order $N^{2}M$, $N_{v}^{2}M$, $NM^{2}$, $N_{v}M^{2}$, $NMN_{v}$, $MN_{v}k$, and $ M^{2}k$.
In the algorithms used to find virtual observation locations, Algorithm \ref{alg:find_xv} and \ref{alg:find_xvi}, we make sure to reuse computations that only
involve the training data in each iteration of $N_{v} = 1, 2, \dots $. This means that in addition to the previously stated operations, we need to 
perform Cholesky factorization of order $N_{v}^{3}$ and generate samples from a $N_{v}$-dimensional truncated Gaussian.
This is initially very cheap, but becomes the main numerical challenge when $N_{v}$ grows large. 
As the purpose of these algorithms is to find a small set $X^{v}$, that also avoids sampling issues due to serial correlation,
we found it useful to output the minimal constraint probability $p^{*}$ found in each iteration to reveal if the stopping criterion 
used (in terms of $p_{target}$ or a maximum number of iterations) was unrealistic in practice. 

\section{Gaussian Process Modelling with Boundedness and Monotonicity Constraints}
\label{sec:monoconstr}

In this section we present some examples related to function estimation where we assume that the function and some of its 
partial derivatives are bounded. This is the scenario considered in the literature on shape-constrained GPs, and alternative 
approaches to GPs under linear constraints are usually presented in this setting. 
We start by a brief discussion on related work, followed by some numerical experiments using boundedness and
monotonicity constraints. The numerical experiments were performed using the 
Python implementation available at \url{https://github.com/cagrell/gp_constr}.

\subsection{Related Work}
\label{sec:related_work}
We give a brief overview of some alternative and related approaches to constrained GPs. 
For the approaches that rely on imposing constraints at a finite set of virtual observation locations, 
we recall that the constraint probability 
can be used in the search for a suitable set of virtual observation locations. The constraint probability 
is the probability that the constraint holds at an arbitrary input $\x$,
$p_{c}(\x)$ given in \eqref{eq:pc}. Some key characteristics of the approaches 
that make use of virtual observations are summarized in Table \ref{tbl:alternatives}.

The related work most similar to the approach presented in this paper is that of \citet{WangBerger:2016:CGP} and \citet{DaVeiga:2012:CGP, DaVeiga:2017:CGP}.
\citet{WangBerger:2016:CGP} make use of a similar sampling scheme for noiseless GP regression applied to computer code emulation. 
A Gibbs sampling procedure is used for inference and to estimate the constraint probability $p_{c}(\x)$ in the search for 
virtual observation locations. The approach of \citet{DaVeiga:2012:CGP, DaVeiga:2017:CGP} is based on 
computation of the posterior mean and covariance of the constrained GP, 
using the equations that are also restated in this paper in Corollary \ref{corr:moments}. They make use of a Genz approximation for inference
\citep{Genz:1992:numerical_comp, Genz:1997:comparisson}, and also 
introduce a crude but faster correlation-free approximation that can be used in the search for 
virtual observation locations. The approach of \citet{DaVeiga:2012:CGP, DaVeiga:2017:CGP} is discussed further in the 
numerical experiments below, where we illustrate the idea in Example 1 and in Example 2 study an approximation 
of the posterior constrained GP using the constrained moments with a Gaussian distribution assumption. 
A major component in \citep{DaVeiga:2012:CGP, DaVeiga:2017:CGP}, \citep{WangBerger:2016:CGP} and this paper is thus 
computation involving the truncated multivariate Gaussian. Besides the choice of method for sampling 
from this distribution, the main difference with our approach 
is that we leverage Cholesky factorizations and noisy virtual observations for 
numerical stability. 

A different approach that also make use of virtual observations is that of \citet{Riihimaki:2010:CGP},
where a \emph{probit} likelihood is used to represent interval observations of the derivative process 
to impose monotonicity. They then make use of Expectation Propagation (EP) to approximate the posterior with a multivariate Gaussian.
As pointed out by \citet{Golchi:2015:CGP}, the Gaussian assumption is questionable if the constraint (in this case monotonicity) 
does not hold with high probability a priori. \citet{Golchi:2015:CGP} proceeds to develop a fully Bayesian procedure 
for application to computer experiments by the use of Sequentially Constrained Monte Carlo Sampling (SCMC).
A challenge with this approach however is that finding a suitable set of virtual observation locations is difficult.
Our experience, in agreement with \citep{WangBerger:2016:CGP, DaVeiga:2012:CGP, DaVeiga:2017:CGP, Riihimaki:2010:CGP},
is that for practical applications in more than a few dimensions, such a strategy is essential to avoid numerical issues related to high serial correlation, and 
also to reduce the number of virtual observation locations needed. 
It is also worth noting that a strategy that decouples computation involving training data and virtual observation locations
from inference at new locations is beneficial. For the approaches discussed herein that rely on sampling/approximation 
related to the truncated multivariate Gaussian, the samples/approximations can be stored and reused as discussed in Section \ref{sec:numerical_cons}.
\begin{table}
    \footnotesize
    \centering
    \begin{tabular}{p{0.225\linewidth}p{0.14\linewidth}p{0.2\linewidth}p{0.28\linewidth}}
        \toprule
         &  \textbf{Virtual obs. likelihood} &  \textbf{Inference strategy} & \textbf{Strategy for finding} $X^{v}$ \\
        \midrule
              Agrell (2019) & Indicator & Sampling & Based on estimating $p_{c}(\x)$ \\
              & + noise & (Minimax tilting) & from samples \\
              & & & \\
              Wang and Berger (2016) & Indicator & Sampling (Gibbs) & Based on estimating $p_{c}(\x)$ \\
              & & & from samples\\
              & & & \\
              Da Veiga and Marrel & Indicator & Moment approxima- & Based on approximating $p_{c}(\x)$ \\
              (2012, 2015) & & tion (Genz) & assuming Gaussian posterior distribution \\
              & & & \\ 
              Riihim{ä}ki and Vehtari & Probit & Expectaion Propaga- & Based on approximating $p_{c}(\x)$ \\
              (2010) & & tion & assuming Gaussian posterior distribution \\
              & & & \\     
              Golchi et al. (2015) & Probit & SCMC & NA \\
        \bottomrule
    \end{tabular}    
    \caption{Summary of alternative approaches that make use of virtual observations. 
    The table compares the likelihood used for virtual observations, the method used for inference and to 
    determine the set of virtual observation locations $X^{v}$.}
    \label{tbl:alternatives}
\end{table}

There are also some approaches to constrained GPs that are not based on the idea 
of using virtual observations. An interesting approach by \citet{Maatouk:2017:CGP}, that is also followed up by \citet{LpezLopera:2017:CGP}, is based on modelling 
a conditional process where the constraints hold in the entire domain.
They achieve this through finite-dimensional approximations of the GP that converge uniformly pathwise. 
With this approach, sampling from a truncated multivariate Gaussian is also needed for inference,
in order to estimate the coefficients of the finite-dimensional approximation that arise 
from discretization of the input space. 
The authors give examples in 1D and 2D, but note that due to the structure of the 
approximation, the approach will be time consuming for practical applications in higher dimensions.
There are also other approaches that consider special types of shape constraints, but where generalization seems difficult. 
See for instance \citep{Abrahamsen:2001:CGP, Yoo:2006:area_to_point, Michalak:2008:Gibbs_ineq, Kleijnen:2013:MonotonicitypreservingBK, Lin:2014:CGP, Lenk:2017:CGP}.

\subsection{Numerical Experiments}
\label{sec:num_exp}
In this section we will make us of the following constraints:

\begin{itemize}
    \item $a_{0}(\x) \leq f(\x) \leq b_{0}(\x)$
    \item $a_{i}(\x) \leq \partial f / \partial x_{i} (\x) \leq b_{i}(\x)$
\end{itemize}
\noindent
for all $\x$ in some bounded subset of $\mathbb{R}^{n_{x}}$, and $i \in \mathcal{I} \subset \{ 1, \dots, n_{x} \}$. Without loss of generality we 
assume that the constrains on partial derivatives are with respect to the first $k$ components of $\x$, 
i.e. $\mathcal{I} = \{1, \dots, k \}$ for some $k \leq n_{x}$. 

As the prior GP we will assume a constant mean $\mu = 0$ and make use of either the RBF or Mat\'ern $5/2$ covariance function. 
These are stationary kernels of the form 
\begin{equation}
    \label{eq:kernel}
    K(\x, \x') = \sigma_{K}^{2} k(r), \text{ } r = \sqrt{ \sum_{i = 1}^{n_{x}} \left( \frac{x_{i} - x_{i}'}{l_{i}} \right)^{2} },
\end{equation}
with variance parameter $\sigma_{K}^{2}$ and length scale parameters $l_{i}$ for $i = 1, \dots, n_{x}$. 
The radial basis function (RBF), also called squared exponential kernel, and the Mat\'ern $5/2$ kernel 
are defined through the function $k(r)$ as 
\begin{equation*}
    k_{\text{RBF}}(r) = e^{-\frac{1}{2}r^{2}} \text{ and } k_{\text{Mat\'ern }5/2}(r) = (1 + \sqrt{5}r + \frac{5}{3}r^{2}) e^{-\sqrt{5}r}.
\end{equation*}
In general, the kernel hyperparameters $\sigma_{K}^{2}$ and $l_{i}$ are optimized together with the noise variance $\sigma$ through MLE.
In the examples that consider noiseless observations, the noise variance is not estimated, but set to a small fixed value as discussed in Section \ref{sec:numerical_cons}.
With the above choice of covariance function, existence of the transformed GP is ensured. In fact, the resulting process
is infinitely differentiable using the RBF kernel \citep[see][Theorem 2.2.2]{Adler:1981:geometry} and 
twice differentiable with the Mat\'ern $5/2$.
These prior GP alternatives were chosen as they are the most commonly used in the literature, 
and thus a good starting point for illustrating the effect of including linear constraints.  
We note that although it is not in general possible to design mean and covariance functions that 
produce GPs that satisfy the constraints considered in this paper, 
one could certainly ease numerical computations by selecting a GP prior 
based on the constraint probability $p(C | Y, \theta)$ in \eqref{eq:constrprob}, 
and for instance make us of a mean function that is known to satisfy the constraint. 

If we let $\sublinop{}^{0}f = f$, $\sublinop{}^{i}f = \partial f / \partial x_{i}$, and  $X^{v, i}$
be the set of $S_{i}$ virtual observations corresponding to the $i$-th operator $\sublinop{}^{i}$, then
we can make use of the formulation in Section \ref{sec:separate_obs} and equations from Appendix \ref{app:search_xv_i} to obtain
\begin{equation*}
    \linop{}\mu^{v} = [\mu \textbf{1}_{S_{0}}, \textbf{0}_{S_{[1, k]}}]^{T},
\end{equation*}
where $\textbf{1}_{S_{1}}$ is the vector $[1, \dots, 1]^{T}$ of length $S_{1}$ and 
$\textbf{0}_{S_{[1, k]}}$ is the vector $[0, \dots, 0]^{T}$ of length $S_{[1, k]} = -S_{0} + \sum S_{i}$.
Furthermore, 
\begin{equation*}
    \begin{aligned}
        K_{X, X^{v}}\linop{}^{T} &= \left[ K^{\ }_{X, X^{v, 0}}, (K^{1,0}_{X^{v, 1}, X})^{T}, \dots, (K^{k,0}_{X^{v, k}, X})^{T} \right] \text{,} \\
        K_{X^{*}, X^{v}}\linop{}^{T} &= \left[ K^{\ }_{X^{*}, X^{v, 0}}, (K^{1,0}_{X^{v, 1}, X^{*}})^{T}, \dots, (K^{k,0}_{X^{v, k}, X^{*}})^{T} \right] \text{,} \\
        \linop{}K_{X^{v}, X^{v}}\linop{}^{T} &=
        \begin{bmatrix*}[c]
            K_{X^{v, 0}, X^{v, 0}} & (K^{1,0}_{X^{v, 1}, X^{v, 0}})^{T} & \dots & (K^{k,0}_{X^{v, k}, X^{v, 0}})^{T} \\
            K^{1,0}_{X^{v, 1}, X^{v, 0}} & K^{1,1}_{X^{v, 1}, X^{v, 1}} & \dots & K^{1,k}_{X^{v, 1}, X^{v, k}} \\
            \vdots & \vdots & \ddots & \vdots \\
            K^{k,0}_{X^{v, k}, X^{v, 0}} & K^{k,1}_{X^{v, k}, X^{v, 1}} & \dots & K^{k,k}_{X^{v, k}, X^{v, k}}
        \end{bmatrix*},     
    \end{aligned}
\end{equation*}
where we have used the notation 
\begin{equation*}
    K^{i, 0}(\x, \x') = \frac{\partial}{\partial x_{i}} K(\x, \x') \text{ and }
    K^{i, j}(\x, \x') = \frac{\partial^{2}}{\partial x_{i} \partial x_{i}'} K(\x, \x').
\end{equation*}
The use of constraints related to boundedness and monotonicity is illustrated using three examples of 
GP regression. Example 1 considers a function $\textnormal{f} : \mathbb{R} \rightarrow \mathbb{R}$ subjected to boundedness
and monotonicity constraints. In Example 2 a function $\textnormal{f} : \mathbb{R}^{4} \rightarrow \mathbb{R}$ is estimated 
under the assumption that information on whether the function is monotone increasing or decreasing as a function 
of the first two inputs is known, i.e. 
$\sgn{\partial \textnormal{f} / \partial x_{1}}$ and $\sgn{\partial \textnormal{f} / \partial x_{2}}$ are known. 
In Example 3 we illustrate how monotonicity constraints in multiple dimensions can be used 
in prediction of pressure capacity of pipelines. 

\subsubsection{\textbf{Example 1}: Illustration of Boundedness and Monotonicity in 1D}
\begin{figure}[H]
    \begin{center}
        \includegraphics[width = 8cm, trim={0.8cm 1.2cm 0.1cm 2.5cm},clip]{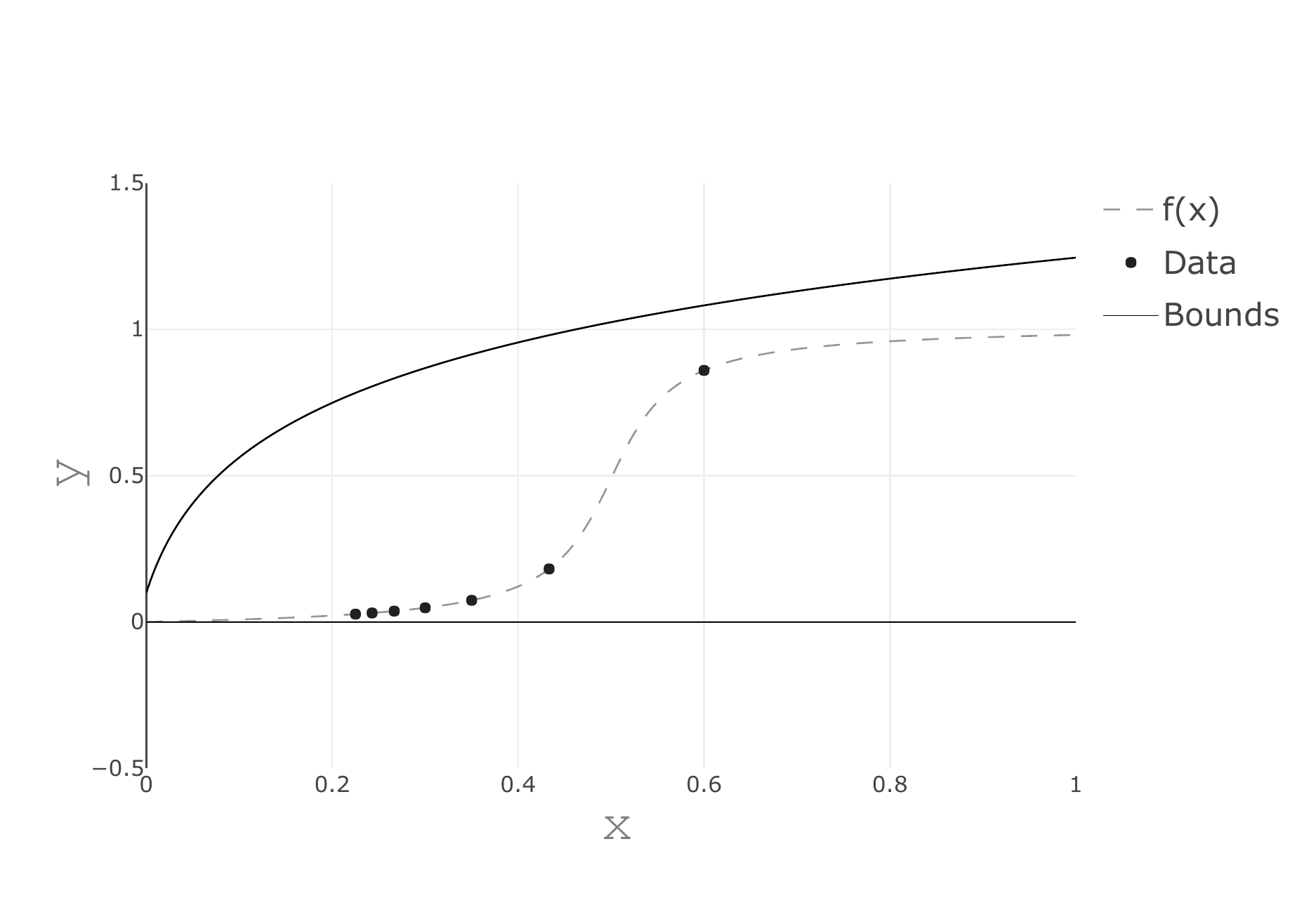}
    \end{center}
    \caption{Function to emulate in Example 1}
    \label{fig:ex1_0}
\end{figure}

As a simple illustration of imposing constraints in GP regression, we first consider the function 
$\textnormal{f} : \mathbb{R} \rightarrow \mathbb{R}$ given by $\textnormal{f}(x) = \frac{1}{3} [\atan{20x - 10} - \atan{-10}]$. 
We assume that the function value is known at $7$ input locations given by 
$x_{i} = 0.1 + 1 / (i+1) $ for $i = 1, \dots, 7$. First, we assume that the observations are noiseless, i.e.
$\textnormal{f}(x_{i})$ is observed for each $x_{i}$. Estimating the function that interpolates at these observations 
is commonly referred to as \textit{emulation}, which is relevant when dealing with data from 
computer experiments. 
Our function $\textnormal{f}(x)$ is both bounded and increasing on all 
of $\mathbb{R}$. In this example we will constrain the GP to satisfy the conditions that for $x \in [0, 1]$, we have that
$\text{d}\textnormal{f} / \text{d}x  \geq 0$ and $a(x) \leq \textnormal{f}(x) \leq b(x)$ for $a(x) = 0$ and $b(x) = \frac{1}{3}\text{ln}(30x + 1) + 0.1$.
The function is shown in Figure \ref{fig:ex1_0} together with the bounds and the $7$ observations.
 
\begin{figure}[H]
    \begin{center}
        \includegraphics[width = 12cm, trim={0.2cm 0.2cm 0.0cm 0.0cm},clip]{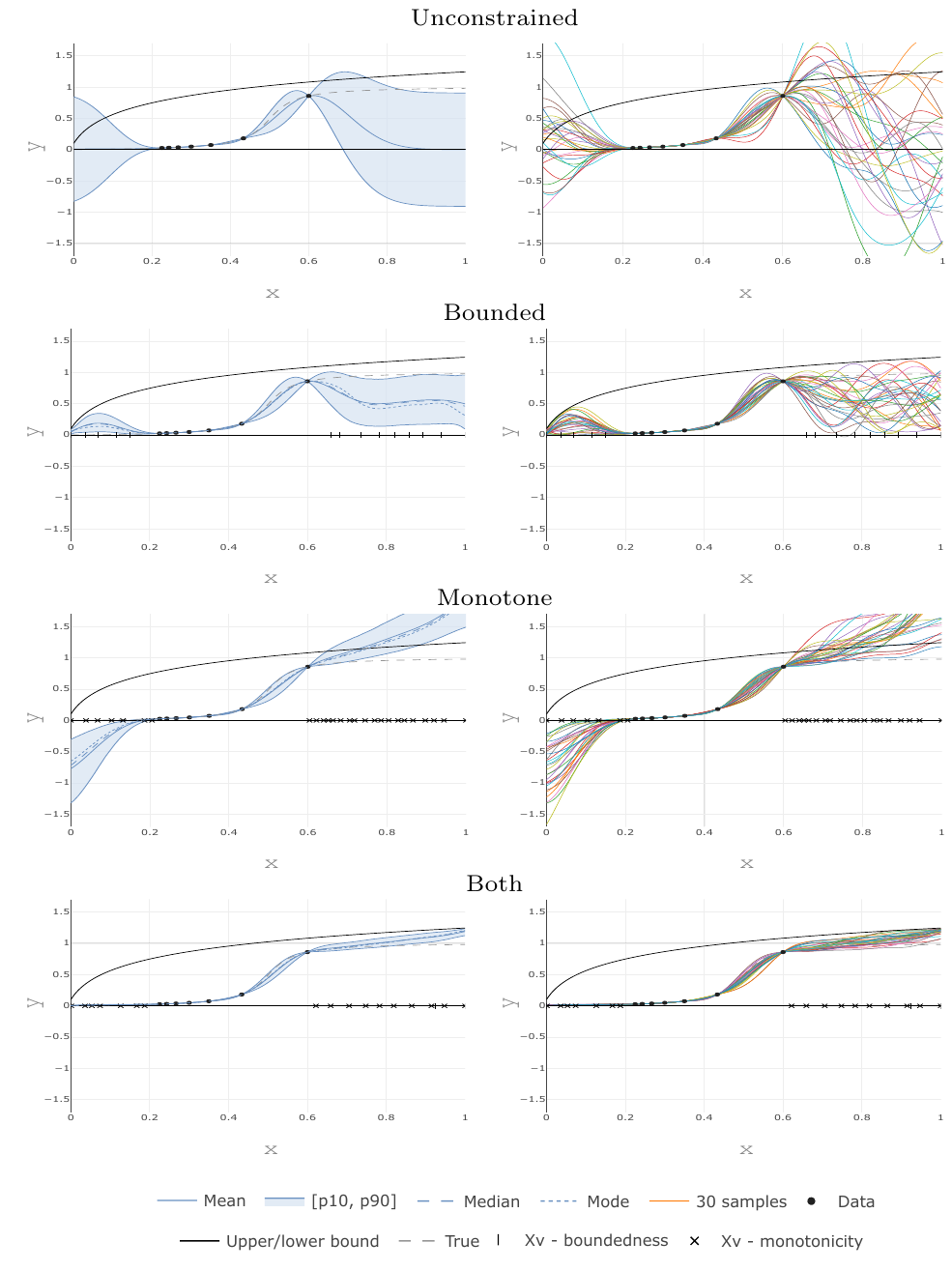}
    \end{center}
    \caption{The GP with parameters $\sigma_{K} = 0.5$ (variance) and $l = 0.1$ (length scale) used in Example 1. 
    The virtual observation locations are indicated by markers on the $x$-axis.}
    \label{fig:ex1_1}
\end{figure}

We select an RBF kernel \eqref{eq:kernel} with parameters $\sigma_{K} = 0.5$ (variance) and $l = 0.1$ (length scale). 
To represent noiseless observations we set $\sigma^{2} = 10^{-6}$, 
where $\sigma^{2}$ is the noise variance in the Gaussian likelihood.
The assumed noise on virtual observations will also be 
set to $10^{-6}$.
To illustrate the effect of 
adding constraints we show the constrained GP using only boundedness constraint, only monotonicity constraint and finally 
when both constraints are imposed simultaneously.
Figure \ref{fig:ex1_1} shows the resulting GPs.
Algorithm \ref{alg:find_xvi} was used with 
a target probability $p_{target} = 0.99$ to determine the virtual observation locations 
that are indicated in the figures,
and the posterior mode was computed by maximizing a Gaussian kernel density estimator 
over the samples generated in Algorithm \ref{alg:sampling}.
For both constraints, $17$ locations was needed for monotonicity 
and only $3$ locations was needed to impose boundedness when the virtual locations for both constraints 
where optimized simultaneously. This is reasonable, as requiring $f(0) > 0$ is sufficient to ensure 
$f(x) > 0$ for $x \geq 0$ when $f$ is increasing, and similarly requiring $f(x^{v}) < b(x^{v})$ 
for some few points $x^{v} \in [0.6, 1]$ should suffice. 
But note that Algorithm \ref{alg:find_xvi} finds the virtual observation locations for both constraints 
simultaneously. Here $x^{v} = 0$ for boundedness was first identified, followed by some few points for monotonicity, 
followed by a new point $x^{v}$ for boundedness etcetera. 

For illustration purposes none of the hyperparameters of the GP were optimized. Moreover, for data sets such as the one in this example
using plug-in estimates obtained from MLE generally not appropriate due to overfitting. Maximizing the marginal likelihood for the unconstrained GP  
gives a very poor model upon visual inspection ($\sigma_{K} = 0.86, l = 0.26$). However, it was observed that the estimated parameters for the constrained model
(using Eq. \eqref{eq:lik_constr}) gives estimates closer to the selected prior which seems more reasonable ($\sigma_{K} = 0.42, l = 0.17$), 
and hence the inclusion of the constraint probability, $p(C | Y, \theta)$, in the likelihood seems to improve the estimates also for the unconstrained GP. 

\begin{figure}[H]
    \begin{center}
        \includegraphics[width = 14cm, trim={0.1cm 0.0cm 0.0cm 0.0cm},clip]{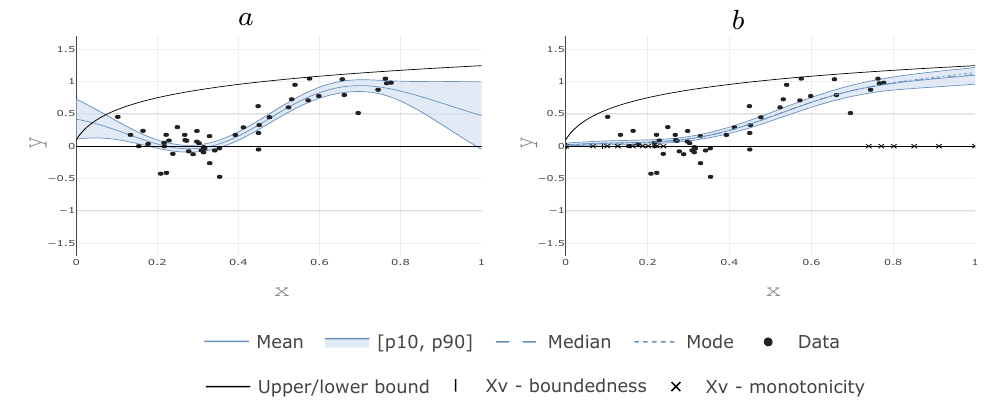}
    \end{center}
    \caption{Unconstrained ($a$) and constrained ($b$) GPs fitted to 50 observations with Gaussian noise.
    The predictive distributions are shown, i.e. the distribution of $f(x)$ where $y = f(x) + \varepsilon$.}
    \label{fig:ex1_2}
\end{figure}

We may also assume that the observations come with Gaussian white noise, which in terms of numerical stability is 
much less challenging than interpolation. Figure \ref{fig:ex1_2} shows the resulting GPs fitted to $50$ observations. 
The observations were generated by sampling $x_{i} \in [0.1, 0.8]$ uniformly, and $y_{i}$ from $\textnormal{f}(x_{i}) + \varepsilon_{i}$
where $\varepsilon_{i}$ are i.i.d. zero mean Gaussian with variance $\sigma^{2} = 0.04$. Both GPs were optimized using plug-in estimates of hyperparameters
($\sigma_{K}, l, \sigma^{2}$) given by maximizing the marginal likelihood. These are ($\sigma_{K} = 0.34, l = 0.32, \sigma^{2} = 0.053$) for the 
constrained case and ($\sigma_{K} = 0.34, l = 0.23, \sigma^{2} = 0.040$) for the unconstrained case. We observe that the estimated noise variance
is larger in the constrained model than the unconstrained where this estimate is exact.

\begin{figure}[H]
    \begin{center}
        \includegraphics[width = 14cm, trim={0.1cm 0.0cm 0.0cm 0.0cm},clip]{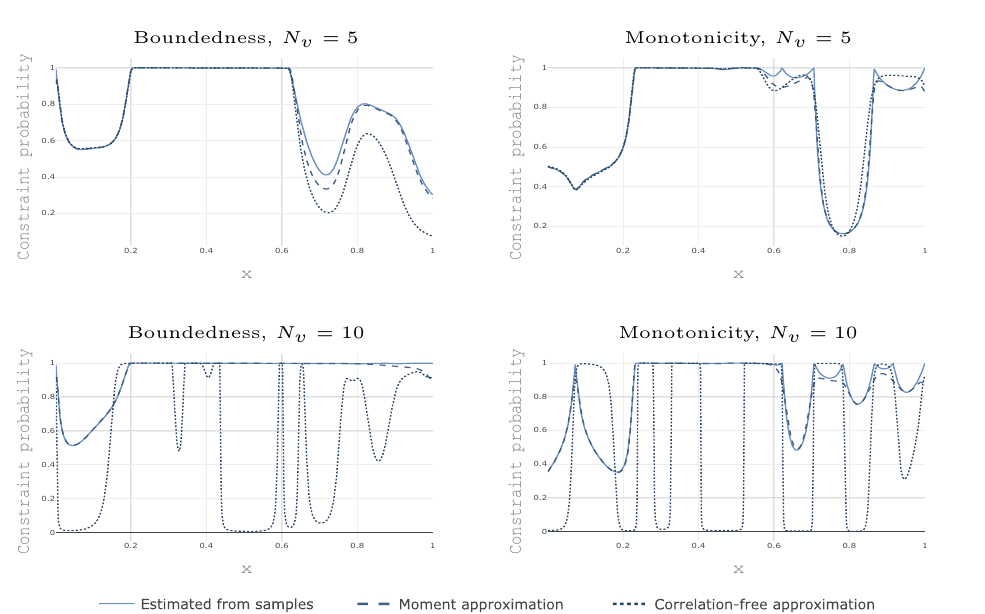}
    \end{center}
    \caption{Constraint probability $p_{c}(\x)$ computed using the estimate \eqref{eq:p_hat_i} 
    together with the moment based approximations from \citet{DaVeiga:2017:CGP}. 
    The constraint probability is shown for monotonicity and boundedness, where 
    $N_{v}$ is the total number of virtual observation locations used in the model. }
    \label{fig:ex1_xv}
\end{figure}

\citet{DaVeiga:2017:CGP} propose to use estimates of the posterior mean and variance of $\linop{}f(\x) | Y, C$
to estimate the constraint probability $p_{c}(\x)$ assuming a Gaussian distribution.  
They also introduce the faster correlation-free approximation, where the parameters are estimated 
under the assumption that observations of $\linop{}f(\x) | Y$ at different input locations $\x$ are independent 
(see Section \ref{sec:moment_approach}).
In Figure \ref{fig:ex1_xv} we plot estimates of $p_{c, i}(\x)$, for the boundedness and monotonicity constraint individually, 
using the approach in this paper \eqref{eq:p_hat_i} 
and the two moment based approximations. The plots were generated first after a total of $5$ and then $10$
virtual observations locations had been included in the model with both constraints. 
As we are mainly interested in finding $\x^{*} = \argmin p_{c, i}(\x)$, Figure \ref{fig:ex1_xv} 
indicates that the moment based approximations are appropriate initially. 
However, as more virtual observation locations are included, the correlation-free assumption becomes questionable. 
But it could still serve as a useful starting point, and in a strategy based on checking the approximation error 
from time to time, it should still be possible 
to take advantage of the computational savings offered by the correlation-free approximation. 

\subsubsection{\textbf{Example 2}: 4D Robot Arm Function}
\label{seq:ex2}
In this example we consider emulation of a function $\textnormal{f} : \mathbb{R}^{4} \rightarrow \mathbb{R}$, where 
we assume that the sign of the first two partial derivatives, $\sgn{\partial \textnormal{f} / \partial x_{1}}$ and $\sgn{\partial \textnormal{f} / \partial x_{2}}$, are known.
The function to emulate is 
\begin{equation*}
    \textnormal{f}(\x) = \sum^{m}_{i=1} L_{i} \ \text{cos} \left( \sum^{i}_{j=1} \tau_{j} \right),
\end{equation*}
for $m = 2$, and $\x = [L_{1}, L_{2}, \tau_{1}, \tau_{2}]$. The function is inspired by the robot arm function often used to 
test function estimation \citep{An:2001:QuasiRegression}. Here $\textnormal{f}(\x)$ is the y-coordinate of a two dimensional robot arm 
with $m$ line segments of length $L_{i} \in [0, 1]$, positioned at angle $\tau_{i} \in [0, 2 \pi]$ with respect to the horizontal axis. 
The constraints on the first two partial derivatives thus implies that it is known whether or not the arm will move further away 
from the x-axis, as a function of the arm lengths, $L_{1}$ and $L_{2}$, for any combination of $\tau_{1}$ and $\tau_{2}$.

In this experiment we first fit an unconstrained GP using $40$ observations taken from a Latin hypercube sample over the input space 
$[0, 1]^{2} \times [0, 2 \pi]^{2}$. A Mat\'ern $5/2$ covariance function is used with plug-in MLE hyperparameters. 
Then, a total of $80$ virtual observation locations are found using the procedure in Algorithm \ref{alg:find_xvi}, where 
we search over a finite candidate set of $1000$ locations in the minimization of the constraint probability. We repeat this procedure 
$100$ times and report performance using the predictivity coefficient $Q^{2}$, predictive variance adequation (PVA)
 and the average width of $95\%$ confidence intervals (AWoCI). 

Given a set of tests $y_{1}, \dots, y_{n_{test}}$ and predictions $\hat{y}_{1}, \dots, \hat{y}_{n_{test}}$, $Q^{2}$ is defined as 
\begin{equation*}
    Q^{2} = 1 - \sum_{i = 1}^{n_{test}}(\hat{y}_{i} - y_{i})^{2} / \sum_{i = 1}^{n_{test}}(\bar{y} - y_{i})^{2},
\end{equation*}
where $\bar{y}$ is the mean of $y_{1}, \dots, y_{n_{test}}$. In our experiments the predictions $\hat{y}_{i}$ are given by 
the posterior mean of the GP. The PVA criterion is defined as 
\begin{equation*}
    \text{PVA} = \left| \text{log} \left( \frac{1}{n_{test}} \sum_{i = 1}^{n_{test}} \frac{(\hat{y}_{i} - y_{i})^{2}}{\hat{\sigma}_{i}^{2}} \right) \right|,
\end{equation*}
where $\hat{\sigma}_{i}^{2}$ is the predictive variance. This criterion evaluates the quality of the predictive variances and to what extent 
confidence intervals are reliable. The smaller the PVA is, the better \citep{Bachoc:2013:CV}. In addition to this criterion, it is also 
useful to evaluate the size of confidence intervals. For this we compute the average width of $95\%$ confidence intervals
\begin{equation*}
    \text{AWoCI} = \frac{1}{n_{test}} \sum_{i = 1}^{n_{test}} (p^{(i)}_{0.975} - p^{(i)}_{0.025}),
\end{equation*}
where $p^{(i)}_{0.975}$ and $p^{(i)}_{0.025}$ are the predicted $97.5\%$ and $2.5\%$ percentiles.

\begin{figure}[H]
    \begin{center}
        \includegraphics[width = 15cm, trim={0.0cm 0.0cm 0.0cm 0.0cm},clip]{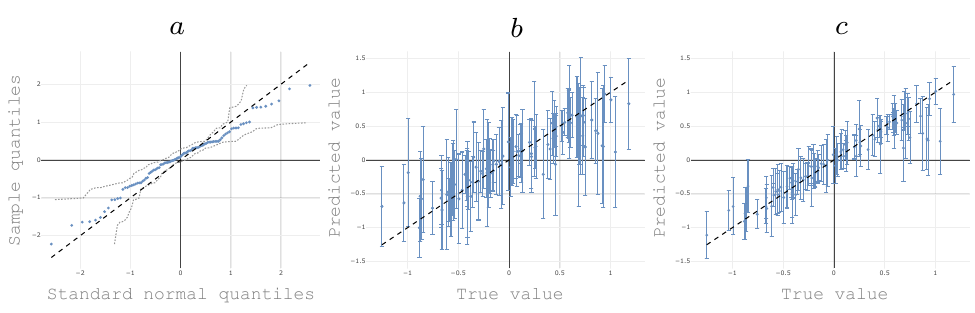}
    \end{center}
    \caption{Figure $a$ shows a qq-plot with $95 \%$ confidence band of $100$ normalized residuals $(y_{i} - \mu_{i})/(\sigma_{i})$, 
    where $\mu_{i}$ and $\sigma_{i}^{2}$ are the mean and variance of the predictive distribution of 
    the unconstrained GP. In Figure $b$, predictions vs the true function value is shown together with a 
    $[0.025, 0.975]$ ($95 \%$) percentile interval for the unconstrained GP. The same type of figure is shown in $c$ for the 
    constrained GP.}
    \label{fig:ex2}
\end{figure}

The result of $100$ predictions for one single experiment is shown in Figure \ref{fig:ex2}. 
As expected, the estimated prediction uncertainty 
is reduced significantly using the constrained model, and single predictions given by the posterior mean 
are also improved. In Table \ref{tbl:robotexp} we summarize the results from running $100$ of these experiments. In each experiment, $Q^{2}$, PVA and AWoCI was 
computed from prediction at $1000$ locations sampled uniformly in the domain. We also report the probability that the constraint holds 
in the unconstrained GP, $p(C|Y)$ given in \eqref{eq:constrprob}, and the CPU time in seconds used to generate $10^{4}$ samples from the posterior 
on an Intel\textsuperscript{\textregistered} Core\textsuperscript{TM} i5-7300U 2.6GHz CPU. 
For comparison, we also include predictions from moment-based approximations using the approach of \citet{DaVeiga:2012:CGP, DaVeiga:2017:CGP}.
We study in particular their approach 
for finding the set of virtual observation locations, as discussed in Section \ref{sec:moment_approach}
and illustrated in the previous example. 
In total, the following alternatives are considered:
\begin{enumerate}
    \item \textbf{Unconstrained}: The initial GP without constraints. 
    \item \textbf{Constrained}: The constrained GP using the approach presented in this paper.
    \item \textbf{Moment approx. 1}: Using the sampling scheme of this paper for inference, but where the 
    moment based approximation is used in the search for virtual observation locations. 
    \item \textbf{Moment approx. 2}: Using moment approximation for both inference and searching for virtual observation locations. 
    This is one of the procedures from \citet{DaVeiga:2012:CGP, DaVeiga:2017:CGP}.
    \item \textbf{Correlation-free approx.}: Same as \textbf{Moment approx. 1} but where 
    the correlation-free approximation is used in the search for virtual observation locations. 
\end{enumerate}

\begin{figure}[H]
    \begin{center}
        \includegraphics[width = 12cm, trim={0.5cm 1.8cm 2cm 2.5cm},clip]{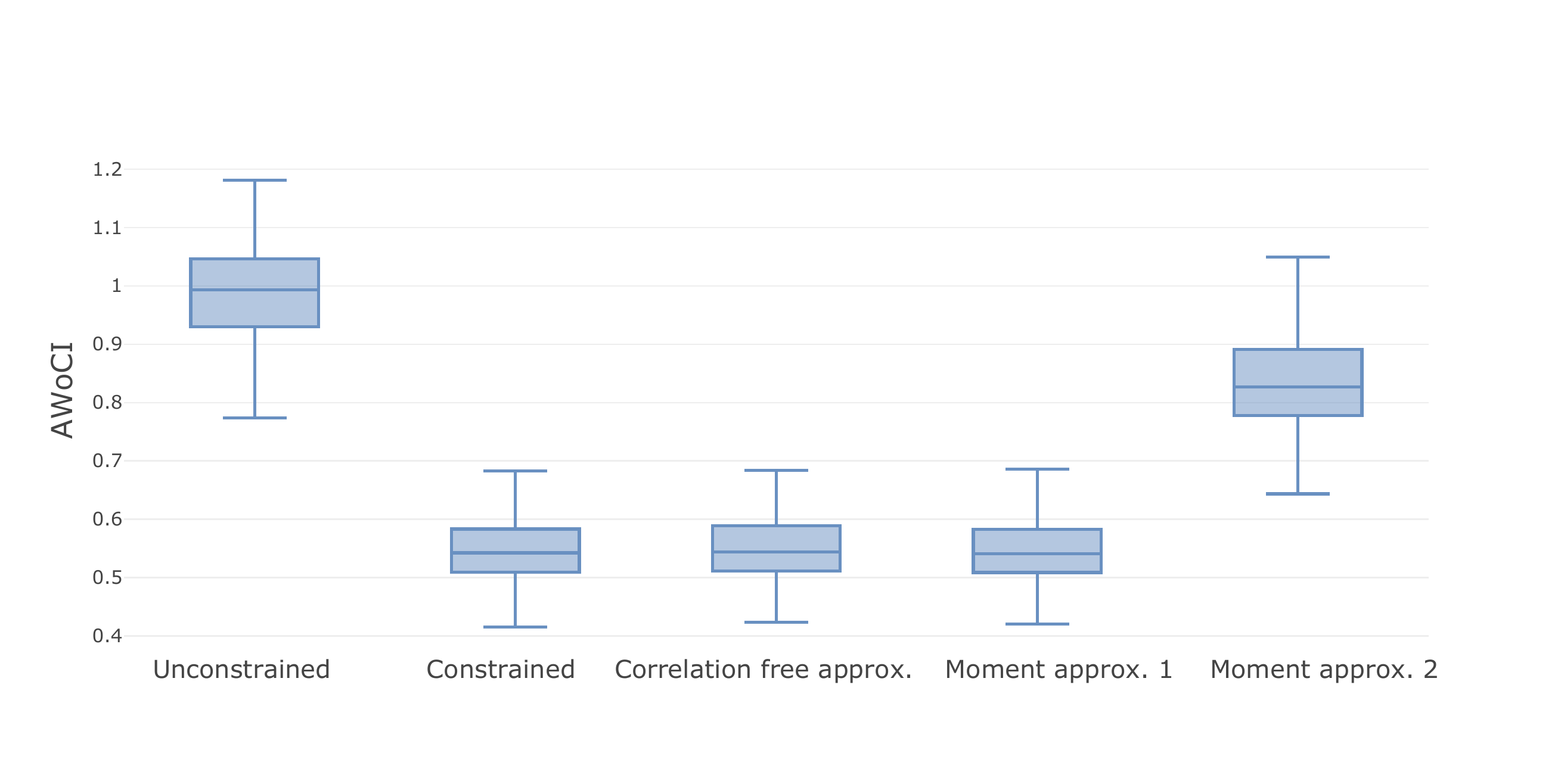}
    \end{center}
    \caption{Average width of confidence intervals (AWoCI) from $100$ experiments of the robot arm function.}
    \label{fig:robot_awoci}
\end{figure}

In Table \ref{tbl:robotexp} we see that the use of constraints is beneficial in terms of both
a higher $Q^{2}$ (better predictive performance) and a smaller PVA (higher quality of predictive variances).
With the exception of 'Moment approx. 2', the inclusion of constraints provides significant uncertainty reduction as 
the width of $95\%$ confidence intervals (AWoCI) are reduce by almost a factor of $2$ on average. 
A box plot showing AWoCI from the $100$ experiments is also shown in Figure \ref{fig:robot_awoci}.
We see that the different approaches for estimating the constraint probability, $p_{c}(\x)$, in the search 
for virtual observation locations work equally well. The Gaussian assumption on the posterior $\textbf{\textup{f}}^{*} | Y, C$
on the other hand is not optimal, as it tends to overestimate the uncertainty in this example. 
\begin{table}
    \centering
    \begin{tabular}{llrrrr}
        \toprule
                     &   $p(C|Y)$ &  $T_{s}$ &   PVA &      $Q^{2}$ &  AWoCI \\
        \midrule
              Unconstrained &       &         &  3.03 &  0.7558 &   0.99 \\
                Constrained &  4.1E-34 &       24.8 &  2.85 &  0.8842 &   0.54 \\
           Moment approx. 1 &  2.4E-36 &       25.2 &  2.84 &  0.8844 &   0.54 \\
           Moment approx. 2 &  2.4E-36 &       25.2 &  2.84 &  0.8844 &   0.83 \\
           correlation-free approx. &  8.6E-37 &       21.1 &  2.91 &  0.8775 &   0.55 \\
        \bottomrule
    \end{tabular}      
    \caption{Average values from $100$ experiments of the robot arm function. 
    $T_{s}$ is the CPU time in seconds used to generate $10^{4}$ samples.}  
    \label{tbl:robotexp}
\end{table}

\subsubsection{\textbf{Example 3}: Pipeline Pressure Capacity}
\label{sec:SRA_ex}
In this example we consider a model for predicting the pressure capacity of a steel pipeline with defects due to corrosion. 
As corrosion is one of the major threats to the integrity of offshore pipelines, experiments are carried out to 
understand how metal loss due to corrosion affects a pipeline's capacity with respect to internal pressure
\citep{Sigurdsson:background_rp_f101, Amaya:2019:Reliability_corroded_pipe_overview}.
These include full scale burst tests and numerical simulation through Finite Element Analysis (FEA).
Results from this type of experiments serve as the basis for current methodologies used in the industry 
for practical assessment of failure probabilities related to pipeline corrosion, such as 
ASME B31G or DNVGL-RP-F101.
We consider experiments related to a single rectangular shaped defect, which is essential to these methodologies. 

To simulate synthetic experiments of the burst capacity of a pipeline with a rectangular defect, we will use
the simplified capacity equation given in in \cite[RP-F101][]{RP-F101}. 
The maximum differential pressure (capacity in MPa) the pipeline can withstand without bursting is 
in the simplified equation given as 
\begin{equation*}
    P_{cap}(\sigma_{u}, D, t, d, l) = 1.05\frac{2t \sigma_{u}}{D - t} \frac{1 - d/t}{1 - \frac{d/t}{Q}} \text{,} \ \ \  Q = \sqrt{1 + 0.31\frac{l^{2}}{Dt}},
\end{equation*}
where $\sigma_{u} \in [450, 550]$ (MPa) is the ultimate tensile strength of the material,
$D \in [10t, 50t]$ (mm) and $t \in [5, 30]$ (mm) are the outer diameter and wall thickness of the pipeline, 
and $d \in [0, t]$ (mm) and $l \in [0, 1000]$ (mm) are the depth and length of the rectangular defect. 

From the physical phenomenon under consideration, we know that 
the capacity of the pipeline will decrease if the size of the 
defect were to increase. 
Similarly, we know that the pipeline capacity increases with 
a higher material strength or wall thickness, and decreases 
as a function of the diameter, all else kept equal. 
In the form of partial derivatives we can express this
information as: $\frac{\partial P_{cap}}{\partial d} < 0$,
$\frac{\partial P_{cap}}{\partial l} < 0$,
$\frac{\partial P_{cap}}{\partial \sigma_{u}} > 0$,
$\frac{\partial P_{cap}}{\partial t} > 0$ and
$\frac{\partial P_{cap}}{\partial D} < 0$.

For convenience we will transform the input variables to the 
unit hypercube.
Let $\textbf{x}$ denote the transformed input vector $\textbf{x} = [x_{1}, \dots, x_{5}]$,
where $x_{1} = (\sigma_{u} - 450)/(550-450)$, $x_{2} = (D/t - 10)/(50 - 10)$, $x_{3} = (t - 5)/(30-5)$, $x_{4} = d/t$ and $x_{5} = l/1000$.  
We will make use of the function 
\begin{equation*}
    f(\textbf{x}) = P_{cap}(\textbf{x}) \text{ for } \x \in [0, 1]^{5},
\end{equation*}
and assume that the burst capacity observed in an experiment 
is $f(\x) + \varepsilon$, where 
$\varepsilon$ is a zero mean Normal random variable with 
variance $\sigma^{2} = 4$. 
The constraints on the partial derivatives after the transformation
becomes: 
$\frac{\partial f}{\partial x_{1}} > 0$,
$\frac{\partial f}{\partial x_{2}} < 0$,
$\frac{\partial f}{\partial x_{3}} > 0$,
$\frac{\partial f}{\partial x_{4}} < 0$ and
$\frac{\partial f}{\partial x_{5}} < 0$
for $\x \in [0, 1]^{5}$.

In this example we thus have five constraints available, represented by bounds on the partial derivative of $f(\x)$ w.r.t. $x_{i}$ for 
$i = 1, \dots, 5$. Besides studying the effect of including all five constraints, we will test some different alternatives using a smaller 
number of constraints, and also lower input dimensions.  
To simulate a lower dimensional version of the capacity equation, we can consider only the fist $n_{x}$ input variables and keep the 
remaining variables fixed. We consider $n_{x} = 3, 4$ and $5$ where we fix $x_{i} = 0.5$ for all $i > n_{x}$. 
For each of these scenarios we will consider $n_{x}$ and $n_{x} - 1$ number of constraints. We let $n_{c}$ denote 
the number of constraints, where using $n_{c}$ constraints means that the bound on $\partial f / \partial x_{i}$ is 
included for $i = 1, \dots, n_{c}$.

In each experiment we start by generating a training set of $N = 5n_{x}$ or $N = 10n_{x}$ LHS samples from $[0, 1]^{n_{x}}$. 
As in the previous example in Section \ref{seq:ex2}, we fit a zero mean GP using a Mat\'ern $5/2$ covariance function and plug-in hyperparameters by MLE. 
We search over a candidate set consisting of $2500$ uniform samples from $[0, 1]^{n_{x}}$ iteratively to update the set 
of virtual observation locations, until the constraint probability at all locations in the candidate set, and for each constraint, is at least $0.7$. 
To check whether this is a reasonable stopping criterion we finish by minimizing the constraint probability for each constraint, 
using the differential evolution \citep{Storn:1997:diffev} global optimization algorithm available in \citep[SciPy][]{scipy}. 

Table \ref{tbl:SRA_res} shows the results for different combinations of input dimensionality $n_{x}$, number of constraints $n_{c}$ 
and number of training samples $N$, where the results in each row is computed from $100$ experiments. 
As in the previous example we report $p(C|Y)$, PVA, $Q^{2}$ and AWoCI, and the CPU time spent generating samples for prediction ($T_{s}$).
We also report the average CPU time used in the search for a new virtual observation location and 
the minimum constraint probability, $p_{c, \text{min}} = \min_{i = 1, \dots n_{c}} \min_{\x \in [0, 1]^{n_{x}}} \hat{p}_{c, i}(\x)$
\eqref{eq:p_hat_i}, computed with differential evolution. Here we make use of $10^{3}$ 
samples to compute the estimate $\hat{p}_{c, i}(\x)$, whereas $10^{4}$ samples are used for the final prediction.
\begin{table}
    \begin{tabular}{rrr|rrlrcccl}
        \toprule
        $n_{x}$ &  $n_{c}$ &  $N$ &  $N_{v}$ &  $T_{v}$ &   $p(C|Y)$ & $p_{c, \text{min}}$ &  $T_{s}$ & PVA & $Q^{2}$ & AWoCI \\
        \midrule
            3 &        2 &     15 &   3.6 &   0.6 &  2.6E-01 &  0.79 &  0.05 &  0.94 (0.89) &  0.95 (0.95) &   3.9 (6.2) \\
            3 &        2 &     30 &   3.5 &   0.6 &  2.5E-01 &  0.78 &  0.04 &  0.89 (0.87) &  0.97 (0.97) &   3.0 (4.8) \\
            3 &        3 &     15 &   5.8 &   0.9 &  1.2E-01 &  0.74 &  0.09 &  1.47 (1.23) &  0.95 (0.95) &   3.7 (6.1) \\
            3 &        3 &     30 &   3.9 &   0.9 &  2.2E-01 &  0.76 &  0.04 &  0.79 (0.79) &  0.97 (0.97) &   3.1 (5.0) \\
            4 &        3 &     20 &  11.8 &   0.9 &  1.5E-02 &  0.67 &  0.19 &  1.40 (1.29) &  0.87 (0.92) &   5.5 (9.4) \\
            4 &        3 &     40 &  11.7 &   0.9 &  6.6E-03 &  0.71 &  0.18 &  0.51 (0.52) &  0.97 (0.97) &   4.1 (6.9) \\
            4 &        4 &     20 &  13.6 &   1.2 &  6.9E-03 &  0.65 &  0.49 &  1.56 (1.31) &  0.91 (0.91) &   5.5 (9.6) \\
            4 &        4 &     40 &  12.8 &   1.2 &  2.7E-03 &  0.69 &  0.19 &  0.50 (0.48) &  0.97 (0.97) &   4.0 (6.7) \\
            5 &        4 &     25 &  14.8 &   1.2 &  6.3E-03 &  0.66 &  0.22 &  1.03 (1.08) &  0.85 (0.83) &  8.3 (14.3) \\
            5 &        4 &     50 &  17.4 &   1.2 &  1.2E-03 &  0.66 &  0.26 &  0.73 (0.78) &  0.90 (0.90) &  6.8 (11.5) \\
            5 &        5 &     25 &  15.5 &   1.5 &  3.1E-03 &  0.65 &  0.24 &  1.12 (1.10) &  0.82 (0.81) &  8.4 (14.4) \\
            5 &        5 &     50 &  20.2 &   1.6 &  1.1E-03 &  0.61 &  0.35 &  0.67 (0.77) &  0.90 (0.90) &  6.5 (11.3) \\
        \bottomrule
    \end{tabular} 
    \caption{Average values from $100$ experiments with input dimensionality $n_{x}$, number of constraints $n_{c}$ 
    and number of training samples $N$. 
    Values in parenthesis correspond to the unconstrained model.
    Here $p_{c, \text{min}}$ is the minimum of the constraint probability for any constraint over the entire domain 
    after a total of $N_{v}$ virtual observation locations have been included. 
    $T_{v}$ is the 
    average CPU time in seconds used to find each of the $N_{v}$ points using $10^{3}$ samples, and 
    $T_{s}$ is the CPU time in seconds used to generate $10^{4}$ samples of the final model for prediction. }  
    \label{tbl:SRA_res}
\end{table}

From Table \ref{tbl:SRA_res} we first notice that the number of virtual observation locations ($N_{v}$) determined 
by the searching algorithm is fairly low. One might interpret this as an indication that the unconstrained GP 
produces samples that are likely to agree with the monotonicity constraints, except for at a few locations. 
As a result, computation that involve sampling from the truncated multivariate Gaussian 
is efficient. 
Still, we see that inclusion of the constraints has an effect on uncertainty estimates as the AWoCI is reduced
by a factor of around $1.6$ in each experiment, whereas PVA and $Q^{2}$ are fairly similar 
for the unconstrained and constrained model overall. 
We also notice that the smallest constraint probability found in the domain using a global optimization technique 
is reduced when the number of constraints or dimensionality is increased. This is expected, as we 
only considered a finite candidate set and not the entire domain 
when searching for the location minimizing the constraint probability. Hence, if we really want to achieve 
a minimal constraint probability larger than $0.7$ in $5$ dimensions, more than $2500$ samples in the candidate set 
would be needed with this strategy, or a global optimizer could be used to identify the remaining virtual 
observation locations needed.   

For the application considered in this example, where uncertainty in the prediction is key to risk assessment, we argue that the effect the constraints have on uncertainty estimates 
makes the inclusion of constraints worthwhile. Modern engineering methodologies that make use of capacity 
predictions as the one illustrated in this example are usually derived in the context of Structural Reliability Analysis (SRA),
where the capacity is combined with a probabilistic representation of load (in this case differential pressure) to 
estimate the probability of failure \citep{madsen2006methods}. 

Alternative methods based on conservative estimates to ensure sufficient safety 
margin between load and capacity are also common. For the application considered herein, this would typically mean 
using a lower percentile instead of the posterior mean in order to represent a conservative capacity. The inclusion of constraints 
can therefore help to avoid unnecessary conservatism due to unphysical scenarios, that are not realistic but 
have positive probability in the unconstrained model. 

Finally, we note that the constraints used in this example are not from differentiating the equation used as stand-in for experiments, but 
from knowledge related to the underlying physical phenomenon. The constraints therefore remain applicable, 
were the experiments to come from physical full-scale tests. This naturally also holds in applications to 
computer code emulation, where we would set the noise term $\varepsilon$ to zero in this example if we 
were to assume that the capacity experiments came from a numerical (FEA) simulation. 
With results from this type of numerical simulation, a noise parameter is usually added to the simulation output as well,
to represent model uncertainty as the numerical simulation is not a perfect representation of the real physical phenomenon. 
Very often the model uncertainty is represented by a univariate Gaussian. An interesting alternative here is 
to instead account for the model uncertainty as observational noise in the GP, where 
the use of constraints may help to obtain a more realistic model uncertainty as well. 

\section{Discussion}
\label{sec:discussion}
The model presented in this paper provides a consistent approach to 
GP regression under multiple linear constraints. The computational 
framework used is based on a sampling scheme which is exact in the limit. 
However, sampling strategies like the one in this paper
can be too numerically demanding as opposed to approximation methods such as 
Laplace approximations, variational Bayesian inference, expectation propagation etcetera.
The choice of using a sampling-based approach came from the author's intended use, which 
relates to machine learning for high-risk and safety-critical engineering applications \citep{Agrell:2018:pitfallml}. 
For these applications, a proper treatment of uncertainty with respect to 
risks and the overall reliability of the system under consideration is essential. 
Making predictions based on past observations in this setting is challenging,
as the consequence of wrong predictions may be catastrophic.  
In addition, critical consequences often relate to infrequent or low probability events, 
where relevant data is naturally scarce. However, there is usually additional knowledge 
available, and today’s methods for assessing risk
tend to rely heavily on understanding the underlying physical
phenomenon. We gave an example in Section \ref{sec:SRA_ex} 
considering prediction of the burst capacity of a pipeline, that may serve as a component 
in a larger model of system reliability. Such models are often graphical, e.g. Bayesian networks,
that are derived from known causal dependencies.  
In this scenario it is essential that the accuracy of numerical estimation- or approximation 
methods can be assessed. In the case where simulation-based methods cannot 
be used due to computational limitations, they still serve as a 
useful benchmark that can help in the development and assessment of suitable approximation-based algorithms.
As for the simulation scheme in this paper, the only computational burden 
lies in sampling from a truncated multivariate Gaussian. 
As this is a fairly general problem, multiple good samplers exist for this purpose.
We found the method of \citet{Botev:2017:minimax_tilting} to work particularly well for 
our applications, as it provides exact sampling in a relevant range of dimensions where 
many alternative sampling schemes fail. Based on a comparison made by \cite{LpezLopera:2017:CGP}, 
we see that the method based on Hamiltonian Monte Carlo by \citet{Pakman:2012:HMC} may 
also be appropriate. 

As we discuss briefly in Section \ref{sec:CGP_MLE}, estimation of hyperparameters 
becomes challenging when the term $p(C|Y, \theta)$ enters the likelihood. 
Moreover, as our approach is based on the use of virtual observation locations, we are 
aware that the task of estimating or optimizing model hyperparameters in general is 
not well defined. This is because the likelihood depends both on the hyperparameters 
and the set of virtual observation locations (Eq. \ref{eq:lik_constr}). 
This problem is neglected in the literature on shape-constrained GPs, 
where it is either assumed that the virtual observation locations are 
known a priori (for low input dimension selecting a space filling sufficiently 
dense design is unproblematic), or the hyperparameters are addressed independently
of these.
To our knowledge the problem of simultaneously estimating hyperparameters and virtual observation 
locations has not yet been addressed. A rather simplistic approach is to iterate between estimating hyperparameter  
and the set of virtual observation locations. However, for higher input dimensions 
this might be problematic altogether, in which case sparse approximations may be needed 
to deal with a large set of virtual observation locations. 
In this setting, it might be more fruitful to view the virtual observation locations as 
additional hyperparameters, in a model approximating the posterior corresponding to 
an sufficiently dense set of virtual observation locations, 
e.g. as in the inducing points framework for scaling GPs to 
large data sets \citep{Matthews:2015:KL}. This is a topic of further research. 

With the approach in this paper, we make use of the probability $p(C | Y)$, 
which is interesting in its own for investigating whether constraints such as e.g. monotonicity
are likely to hold given a set of observations. Alternatively, inference on the constraint
noise parameter $\sigma_{v}$ can provide similar type of information. 
Ideally, we choose a small fixed value for $\sigma_{v}$ to avoid numerical instability, 
as discussed in Section \ref{sec:numerical_cons}. But in extreme cases, with conflicting constraints or 
observations that contradict constraints with high probability, the model may still 
experience numerical issues. 
We argue that models that 'break' under these circumstances 
are preferred as it reveals that either 1) there is something wrong with the observations, 
or 2) there is something wrong with the constraints and hence our knowledge of the underlying phenomenon
\citep{Agrell:2018:pitfallml}. It would nevertheless be better if more principled ways of investigating such issues 
were available. 
In our experiments we observed that the conditional likelihood, $p(Y|C)$, in general is 
decreasing as a function of $\sigma_{v}$, whereas this was not the case for an invalid constraint 
assuming a monotonic \emph{decreasing} function in Example 1. Hence, $\sigma_{v}$ 
might provide useful information in this manner. 
The estimated partial constraint probabilities $\hat{p}_{c, i}(\x)$ can also 
be useful for revealing such issues, for instance by monitoring the intermediate minimum values 
$p^{*}_{i}$ computed in Algorithm \ref{alg:find_xvi} as new virtual observation locations are added.

Finally, we note that as the model presented in this paper relies on conditioning on a transformed GP with 
values in $\mathbb{R}^{n_{c}}$, it could be extended to multi-output GPs over functions $\textnormal{f} : \mathbb{R}^{n_{x}} \rightarrow \mathbb{R}^{n_{y}}$ in a natural way. 
But for non-Gaussian likelihoods, or applications with large or high-dimensional data, other approximation based 
alternatives are needed. 

\acks{
This work has been supported by grant 276282 from the Norwegian Research Council and 
DNV GL Group Technology and Research. The research is part of an initiative on applying 
constraints based on phenomenological knowledge in probabilistic machine learning for 
high-risk applications, and the author would like to thank colleagues at DNV GL and 
the University of Oslo for fruitful discussions on the topic. 
A special thanks to Arne B. Huseby, Simen Eldevik, Andreas Hafver, and the 
editor and reviewers of JMLR for insightfull comments that have greatly 
improved the paper.
}

\renewcommand{\theHsection}{A\arabic{section}} 
\appendix
\section{Proof of Lemma \ref{lemma:posterior}}
\label{app:proof:posterior}
{\bf Proof}. We start by observing that $(\f^{*},\widetilde{C}, Y)$ is jointly Gaussian with mean and covariance 
\begin{align}
    \mathbb{E}([\f^{*},\widetilde{C}, Y]^{T}) &= [\mu^{*}, \linop{} \mu^{v}, \mu]^{T}, \label{eq:app_joint_mean} \\
    \text{cov}([\f^{*},\widetilde{C}, Y]^{T}) &=
    \begin{bmatrix*}[c]
        K_{X^{*}, X^{*}} &  K_{X^{*}, X^{v}} \linop{}^{T} & K_{X^{*}, X}  \\
        \linop{} K_{X^{v}, X^{*}} & \linop{}  K_{X^{v}, X^{v}} \linop{}^{T} + \sigma^{2}_{v} I_{N_{v}} & \linop{} K_{X^{v}, X}\\
        K_{X, X^{*}} & K_{X, X^{v}} \linop{}^{T} &  K_{X, X} + \sigma^{2}I_{N}
    \end{bmatrix*}. \label{eq:app_joint_cov}
\end{align}
By first conditioning on $Y$ we obtain
\begin{equation}
    \left.
    \begin{matrix*}[c]
        \f^{*} \\
        \widetilde{C} \\
    \end{matrix*}
    \ \right|
    Y
    \sim \mathcal{N} 
    \left( 
        \begin{bmatrix*}[c]
            \mu^{*} + A_{2}(Y - \mu) \\
            \linop{} \mu^{v} + A_{1}(Y - \mu) \\
        \end{bmatrix*}
        ,
        \begin{bmatrix*}[c]
            B_{2} & B_{3} \\
            B_{3}^{T} & B_{1} \\
        \end{bmatrix*}
    \right),
\end{equation}
for $A_{1} = (\linop{} K_{X^{v}, X})(K_{X, X} + \sigma^{2}I_{N})^{-1}$,
$A_{2} = K_{X^{*}, X}(K_{X, X} + \sigma^{2}I_{N})^{-1}$,
$B_{1} = \linop{} K_{X^{v}, X^{v}} \linop{}^{T} + \sigma^{2}_{v} I_{N_{v}} - A_{1} K_{X, X^{v}} \linop{}^{T}$,
$B_{2} = K_{X^{*}, X^{*}} - A_{2} K_{X, X^{*}}$,
and $B_{3} = K_{X^{*}, X^{v}} \linop{}^{T} - A_{2} K_{X, X^{v}}\linop{}^{T}$.\\

Conditioning on $\widetilde{C}$ then gives
\begin{equation}
    \label{eq:app_posterior_f}
    \f^{*} | Y, \widetilde{C} \sim \mathcal{N} 
    \left( 
        \mu^{*} + A(\widetilde{C} - \linop{}\mu^{v}) + B(Y - \mu), \Sigma
    \right),
\end{equation}
for 
$A = B_{3}B_{1}^{-1}$, 
$B = A_{2} - AA_{1}$
and $\Sigma = B_{2} - AB_{3}^{T}$. 
\\

Similarly, we may derive $\widetilde{C} | Y$ by observing that
the joint distribution of $\widetilde{C}, Y$ is given by removing the first row in \eqref{eq:app_joint_mean} and
the first row and column in \eqref{eq:app_joint_cov}. Hence,
\begin{equation}
    \label{eq:app_posterior_C}
    \widetilde{C} | Y \sim \mathcal{N} 
    \left( 
        \linop{}\mu^{v} + A_{1}(Y - \mu), B_{1}
    \right).
\end{equation}
The constrained posterior of $\widetilde{C}$ is obtained by applying the constraint $C$ to the posterior, 
and hence $\widetilde{C} | Y, C$ becomes a truncated Gaussian with the same mean and variance as in \eqref{eq:app_posterior_C},
and the bounds $a(X^{v})$ and $b(X^{v})$ given by $C$. Similarly, $\f^{*} | Y, C$ is obtained by replacing $\widetilde{C}$ in 
\eqref{eq:app_posterior_f} with $\widetilde{C} | Y, C$. Finally, the probability $p(C | Y)$ is just the probability that $\widetilde{C} | Y$
given in \eqref{eq:app_posterior_C} falls within the bounds given by $C$, 
and the unconstrained distribution remains the same as \eqref{eq:standard_gp_posterior}.
\QEDS

\section{Proof of Lemma \ref{lemma:posterior_chol}}
\label{app:proof:chol}
{\bf Proof}. The equations in Lemma \ref{lemma:posterior_chol} can be verified by simply inserting $L$, $v_{1}$ and $v_{2}$ and 
check against the expressions in Lemma \ref{lemma:posterior}. We show this for $A_{1}$ and $B_{1}$, and the 
results for the remaining matrices are proved by applying the same procedures. In order to factorize $B_{1}$,
we use that $B_{1}$ is the covariance matrix of a Gaussian random variable (see Equation \ref{eq:app_posterior_C} in Appendix \ref{app:proof:posterior}),
and must therefore be symmetric and positive definite. 
\\

To show that $A_{1} = (L^{T} \setminus v_{1})^{T}$ we use that $v_{1} = L \setminus  K_{X, X^{v}} \linop{}^{T}
\Rightarrow Lv_{1} =  K_{X, X^{v}} \linop{}^{T}$. Hence,
\begin{equation*}
    \begin{split}
        &A_{1} = (L^{T} \setminus v_{1})^{T} \\
        &\Rightarrow L^{T}A_{1}^{T} = v_{1} = L \setminus  K_{X, X^{v}} \linop{}^{T} \\ 
        &\Rightarrow LL^{T}A_{1}^{T} = K_{X, X^{v}} \linop{}^{T} \\ 
        &\Rightarrow A_{1} = ((LL^{T})^{-1}  K_{X, X^{v}}\linop{}^{T})^{T} = (\linop{} K_{X^{v}, X})(K_{X, X} + \sigma^{2}I_{N})^{-1},
    \end{split}
\end{equation*}
where we have used that $( K_{X, X^{v}} \linop{}^{T})^{T} = \linop{} K_{X^{v}, X}$
and $LL^{T} = K_{X, X} + \sigma^{2}I_{N}$.
\\

To show that $B_{1} = \linop{} K_{X^{v}, X^{v}} \linop{}^{T} + \sigma^{2}_{v} I_{N_{v}} - v_{1}^{T}v_{1}$
we need to show that $v_{1}^{T}v_{1} = A_{1} K_{X, X^{v}} \linop{}^{T}$, which is trivial
\begin{equation*}
    \begin{split}
        v_{1}^{T}v_{1} &= (L^{-1} K_{X, X^{v}} \linop{}^{T})^{T}(L^{-1} K_{X, X^{v}}\linop{}^{T}) \\   
        &= \linop{} K_{X^{v}, X}(LL^{T})^{-1}  K_{X, X^{v}} \linop{}^{T} \\
        & = A_{1} K_{X, X^{v}} \linop{}^{T}.
    \end{split}
\end{equation*}

\QEDS

\section{Algorithm for Finding Virtual Observation Locations based on Individual Sub-operators}
\label{app:search_xv_i}
We present the details of the algorithm for finding virtual observation locations introduced in Section \ref{sec:separate_obs}.
Here we let $\linop{}$ be a linear operator defined by the column vector $[\sublinop{1}, \dots, \sublinop{k}]$, 
where $\sublinop{i}$ produces functions from $\mathbb{R}^{n_{x}}$ to $\mathbb{R}$, subjected to an interval constraint $[a_{i}(\x), b_{i}(\x)]$. 
We would like to impose constraints related to the i-th sub-operator only at locations where $p(\sublinop{i}f(\x) \notin [a_{i}(\x), b_{i}(\x)])$ 
is not sufficiently small.
For this we let $X^{v}$ be the concatenation of the matrices $X^{v, 1}, \dots, X^{v, k}$
and define $\linop{}^{T}f(X^{v}) = [\sublinop{1}^{T}f(X^{v, 1}), \dots, \sublinop{k}^{T}f(X^{v, 1})]^{T}$.
The matrices needed to make use of Lemma \ref{lemma:posterior} and Lemma \ref{lemma:posterior_chol} are
$\linop{}\mu^{v}$,
$K_{X, X^{v}} \linop{}^{T}$,
$K_{X^{*}, X^{v}} \linop{}^{T}$,
and $\linop{}K_{X^{v}, X^{v}} \linop{}^{T}$. 
Using that $\sublinop{i}f(X^{v}) = \sublinop{i}f(X^{v, i})$, these are given by
\begin{equation*}
    \begin{array}{ll}
        \linop{}\mu^{v} = 
        \begin{bmatrix*}[c]
            \sublinop{1}\mu(X^{v, 1}) \\
            \vdots \\
            \sublinop{k}\mu(X^{v, k}) \\
        \end{bmatrix*}, & 
        K_{X, X^{v}} \linop{}^{T} = 
        \begin{bmatrix*}[c]
            K_{X, X^{v, 1}} \sublinop{1}^{T} \\
            \vdots \\
            K_{X, X^{v, k}} \sublinop{k}^{T} \\
        \end{bmatrix*},
    \end{array}        
\end{equation*}
where $K_{X^{*}, X^{v}} \linop{}^{T}$ also is given by the above equation for $X = X^{*}$.
Finally, $\linop{}K_{X^{v}, X^{v}} \linop{}^{T}$ is the block matrix with blocks
\begin{equation*}
    (\linop{}K_{X^{v}, X^{v}}\linop{}^{T})_{i,j} = \sublinop{i} K_{X^{v, i}, X^{v, j}} \sublinop{j}^{T}.
\end{equation*}
We want to improve the algorithm in Section \ref{sec:find_xv} for finding the set of virtual observation locations 
by considering each sub-operator individually. To do this we make use estimated partial constraint probabilities 
(given in \eqref{eq:p_hat_i} and restated below).
\begin{equation*}
    \hat{p}_{c, i}(\x) = \frac{1}{m} \sum_{j = 1}^{m} P \left( a_{i}(\x) - \nu < (\linop{}f(\x) | Y, C_{j})_{i} < b_{i}(\x) + \nu \right),
\end{equation*}
where $(\linop{}f(\x) | Y, C_{j})_{i}$ is the univariate Normal distribution given by the $i$-th row of $(\linop{}f(\x) | Y, C_{j})$
and $C_{1}, \dots, C_{m}$ are $m$ samples of $\textbf{C}$ given in \eqref{eq:posterior_2} as before.     
For the individual sub-operators $\sublinop{i}$, the set of virtual observations $X^{v}_{i}$ needed to ensure that $\hat{p}_{c, i}(\x) \geq p_{\text{target}}$
can then be found using the following algorithm. 

\begin{alg}
    \label{alg:find_xvi}
    Finding locations of virtual observations $X^{v}_{i}$ s.t. $\hat{p}_{c, i}(\x) \geq p_{\text{target}}$ for all $\x \in \Omega$
    and all sub-operators $\sublinop{1}, \dots, \sublinop{k}$.
    \begin{enumerate}
        \item Compute $L = Chol(K_{X, X} + \sigma^{2}I_{N})$.
        \item Until convergence do:
        \begin{enumerate}
            \item If $X^{v} \neq \emptyset$ compute $A_{1}$ and $B_{1}$ as defined in Lemma \ref{lemma:posterior_chol},
            and generate $m$ samples $C_{1}, \dots, C_{m}$ of $\textbf{C}$ given in \eqref{eq:posterior_2}.
            \item If $X^{v} = \emptyset$ compute $(\x^{*}_{i}, p^{*}_{i}) = (\argmin p_{c, i}(\x), p_{c, i}(\x^{*}))$. Otherwise compute 
            $(\x^{*}_{i}, p^{*}_{i}) = (\argmin \hat{p}_{c, i}(\x), \hat{p}_{c, i}(\x^{*}))$, 
            for all $i = 1, \dots, k$ with $\hat{p}_{c, i}$ defined as in 
            \eqref{eq:p_hat_i} using the samples generated in step (a).
            \item Let $(\x^{*}, p^{*}, j)$ correspond to the smallest probability: 
            $p^{*} = p^{*}_{j} = \text{min}_{i} \ p^{*}_{i}$.
            \item Terminate if $p^{*} \geq p_{\text{target}}$, 
            otherwise update $X^{v}_{j} \rightarrow X^{v}_{j} \cup \{ \x^{*} \}$.
        \end{enumerate}
    \end{enumerate}

\end{alg}

\section{Proof of Lemma \ref{lemma:constr_posterior}}
\label{app:proof:constr_posterior}
{\bf Proof}. This follows exactly from the proofs of Lemma \ref{lemma:posterior}
and Lemma \ref{lemma:posterior_chol} by replacing $\f^{*} \rightarrow \linop{}f(\x^{*})$, 
which implies
$\mu^{*} \rightarrow \linop{}\mu^{*}$, 
$K_{X^{*}, X} \rightarrow \linop{}K_{\x^{*}, X}$,
$K_{X^{*}, X^{*}} \rightarrow \linop{}K_{\x^{*}, \x^{*}} \linop{}^{T}$
and $K_{X^{*}, X^{v}}\linop{}^{T} \rightarrow \linop{}K_{\x^{*}, X^{v}}\linop{}^{T}$.

\QEDS

\section{Proof of Corollary \ref{corr:moments}}
\label{app:proof:corr:moments}
{\bf Proof}. We show the derivation of the expectation and covariance of $\f^{*} | Y, C$ as the 
derivations for $\linop{}f(\x^{*}) | Y, C$ are equivalent. From Lemma \ref{lemma:posterior} we have that
\begin{equation*}
    \textbf{\textup{f}}^{*} | Y, C \sim \mathcal{N}(\mu^{*} + A(\textbf{C} - \linop{} \mu^{v}) + B(Y - \mu), \Sigma ).
\end{equation*}
If we let $\nu, \Gamma$ be the expectation and covariance of $\textbf{C}$, then
\begin{equation*}
    \begin{split}
        \mathbb{E}[\f^{*} | Y, C] = \mathbb{E}_{\textbf{C}}\left[ \mathbb{E}[\f^{*} | Y, \textbf{C}] \right]
        &= \mathbb{E}_{\textbf{C}}\left[ \mu^{*} + A(\textbf{C} - \linop{} \mu^{v}) + B(Y - \mu) \right] \\
        &= \mu^{*} + A(\nu - \linop{} \mu^{v}) + B(Y - \mu), 
    \end{split}
\end{equation*}
and
\begin{equation*}
    \begin{split}
        \text{cov}[\f^{*} | Y, C] &= \mathbb{E}_{\textbf{C}}\left[ \text{cov}[\f^{*} | Y, \textbf{C}] \right] +  \text{cov}_{\textbf{C}}[\mathbb{E}[\f^{*} | Y, \textbf{C}]] \\
        &= \mathbb{E}_{\textbf{C}}[\Sigma] +  \text{cov}_{\textbf{C}}[\mu^{*} + A(\textbf{C} - \linop{} \mu^{v}) + B(Y - \mu)] \\
        &= \Sigma + \text{cov}_{\textbf{C}}[A\textbf{C}] = \Sigma + A \Gamma A^{T}.
    \end{split}
\end{equation*}

\QEDS

\vskip 0.2in
\bibliography{19-065}

\end{document}